\pdfoutput=1

\documentclass[11pt]{article}

\usepackage[final]{acl}

\usepackage{times}
\usepackage{latexsym}

\usepackage[T1]{fontenc}

\usepackage[utf8]{inputenc}

\usepackage{microtype}

\usepackage{inconsolata}

\usepackage{graphicx}

\usepackage{amsmath} 
\usepackage{amssymb} 
\usepackage{tabularx}
\usepackage{booktabs} 
\usepackage[most]{tcolorbox} 
\usepackage{multicol}
\usepackage{textcomp} 
\usepackage{xcolor} 
\definecolor{myblue}{RGB}{70, 130, 180}
\definecolor{tabelcellgrey}{HTML}{f4f5f5} 
\usepackage{makecell} 
\usepackage{colortbl} 
\usepackage{multirow}
\usepackage{rotating} 
\usepackage{adjustbox} 
\usepackage{geometry} 
\usepackage{soul} 
\usepackage{tikz} 
\usepackage{algorithm} 
\usepackage{algpseudocode} 
\usetikzlibrary{shapes.geometric, arrows} 

\definecolor{Magenta}{rgb}{0.8, 0.1, 0.6}

\title{Evaluating Multimodal Large Language Models on Video Captioning \\ via Monte Carlo Tree Search}

\author{
 \textbf{Linhao Yu\textsuperscript{\rm{1}}},
 \textbf{Xinguang Ji\textsuperscript{\rm{2}}},
 \textbf{Yahui Liu\textsuperscript{\rm{2}}},
 \textbf{Fanheng Kong\textsuperscript{\rm{2}}},
 \textbf{Chenxi Sun\textsuperscript{\rm{2}}},
\\
 \textbf{Jingyuan Zhang\textsuperscript{\rm{2}}},
 \textbf{Hongzhi Zhang\textsuperscript{\rm{2}}},
 \textbf{V. W.\textsuperscript{\rm{2}}},
 \textbf{Fuzheng Zhang\textsuperscript{\rm{2}}},
 \textbf{Deyi Xiong\textsuperscript{\rm{1}}}\thanks{Corresponding author.}
\\
 \textsuperscript{1}TJUNLP Lab, College of Intelligence and Computing, Tianjin University, Tianjin, China \\
 \textsuperscript{2}Independent Researcher
\\
\texttt{linhaoyu@tju.edu.cn, dyxiong@tju.edu.cn}
}

\begin{document}
\maketitle

\begin{abstract}

Video captioning can be used to assess the video understanding capabilities of Multimodal Large Language Models (MLLMs).  
However, existing benchmarks and evaluation protocols suffer from crucial issues, such as inadequate or homogeneous creation of key points, exorbitant cost of data creation, and limited evaluation scopes.
To address these issues, we propose an automatic framework, named AutoCaption, which leverages Monte Carlo Tree Search (MCTS) to construct numerous and diverse descriptive sentences (\textit{i.e.}, key points) that thoroughly represent video content in an iterative way.
This iterative captioning strategy enables the continuous enhancement of video details such as actions, objects' attributes, environment details, etc. We apply AutoCaption to curate MCTS-VCB, a fine-grained video caption benchmark covering video details, thereby enabling a comprehensive evaluation of MLLMs on the video captioning task.
We evaluate more than 20 open- and closed-source MLLMs of varying sizes on MCTS-VCB. Results show that MCTS-VCB can effectively and comprehensively evaluate the video captioning capability, with Gemini-1.5-Pro achieving the highest F1 score of 71.2.
Interestingly, we fine-tune InternVL2.5-8B with the AutoCaption-generated data, which helps the model achieve an overall improvement of 25.0\% on MCTS-VCB and 16.3\% on DREAM-1K, further demonstrating the effectiveness of AutoCaption. 
The code and data are available at \url{https://github.com/tjunlp-lab/MCTS-VCB}.

\end{abstract}
\section{Introduction}
\label{sec:introdcution}

\begin{figure}[ht]
    \centering
    \includegraphics[width=\linewidth]{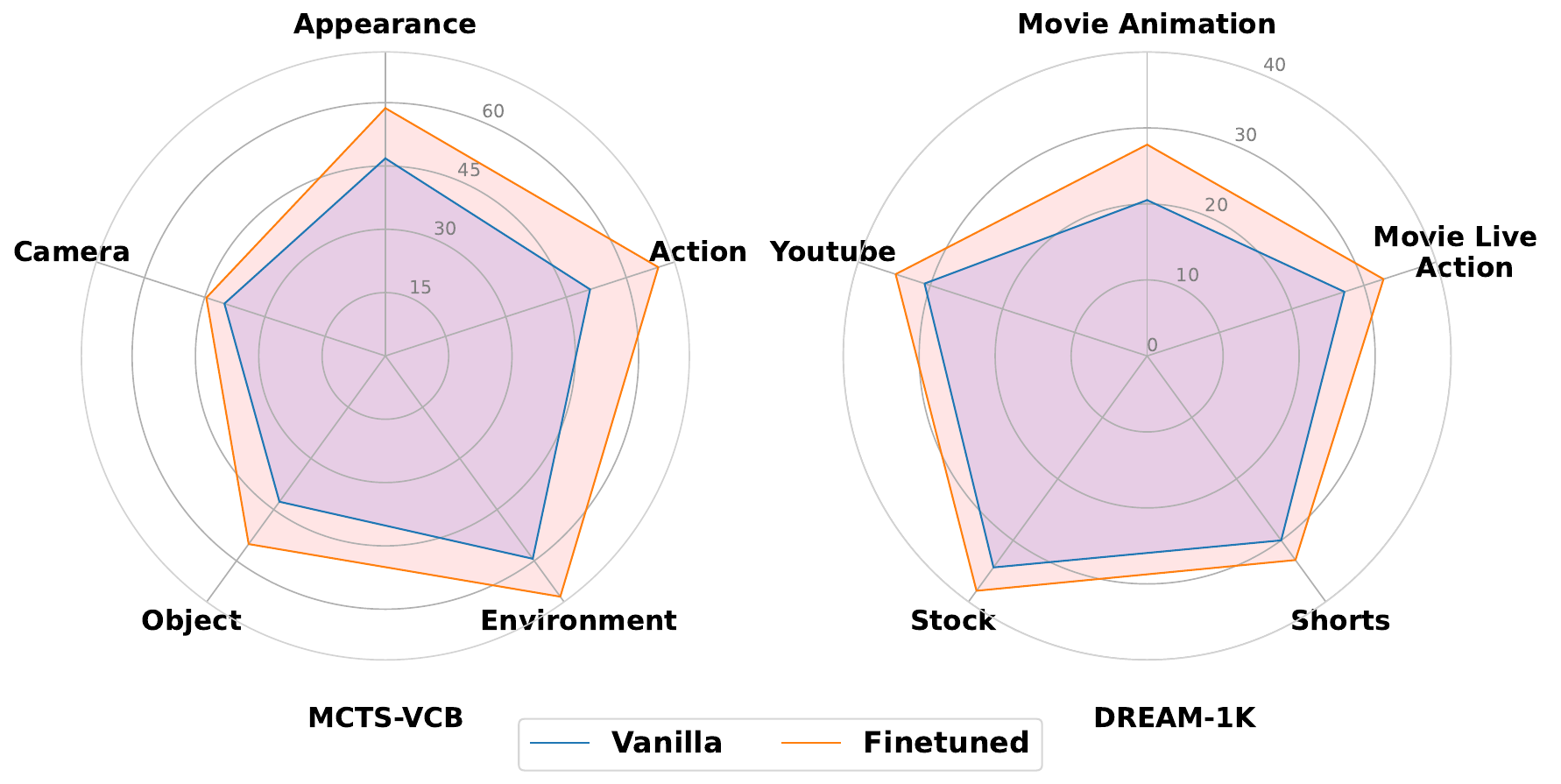}
    \caption{Comparison of MCTS-VCB against DREAM-1K with InternVL2.5-8B~\cite{chen2024internvl} after fine-tuning on caption data generated by AutoCaption.}
    \label{fig:sft_performance_radio}
\end{figure}


Video captioning is a crucial task that expects to obtain a complete, accurate, and low-hallucination textual description for a given video.
Recently, this task is widely applied to showcase the multidimensional and comprehensive understanding capabilities of MLLMs across various video categories~\cite{DBLP:journals/corr/abs-2409-12191,li2024llamavid,hong2025motionbench}.

Previous video captioning benchmarks~\cite{DBLP:journals/corr/abs-2407-00634, DBLP:conf/acl/ChenD11, DBLP:conf/cvpr/XuMYR16, DBLP:conf/iccv/WangWCLWW19} commonly involve creating key points (\textit{i.e.}, descriptive sentences) from video clips and comparing them to the captions generated by MLLMs.
There are \textit{three} main challenges by using the above strategy: 1) Once the created key points are inadequate or homogeneous, the evaluation can be overstated or inaccurate. For example, if the created key points are tend to general or action-oriented descriptions, they would not be suitable for evaluating the perception of environment aspects. 2) Owing to the potential incompleteness of annotations, it is challenging to perform a full-scope evaluation of the video captioning capability of MLLMs. 3) Either manually annotated captions or  extracted from manually constructed captions are time-consuming and expensive.
To alleviate the above-mentioned challenges, we introduce AutoCaption, an automated framework that utilizes Monte Carlo Tree Search (MCTS) to iteratively construct plenty of diverse key points, corresponding to the given video content. 
AutoCaption can enrich the description of video details (\textit{e.g.}, color, quantity, position, material, and other attributes of objects) through a continuous procedure of predefined actions using a tree search approach. In this manner, AutoCaption can eliminate the need for labor-intensive manual annotations, and address the challenge of creating insufficient or homogeneous key points.

Based on AutoCaption, we have established a video caption benchmark, named MCTS-VCB, which can be applied to assess the video captioning capabilities of MLLMs. Our MCTS-VCB benchmark consists of 1,765 videos that are sourced from 10 distinct video categories. Each video contains an average of 122 key points that are categorized into five dimensions, involving a broad evaluation scope.

After evaluating 20+ open-source and closed-source 
MLLMs on MCTS-VCB, we observe that the performance of different MLLMs varies a lot, especially on some particular video categories.
To further validate the effectiveness of AutoCaption, we construct a dataset of around 10K samples with AutoCaption to fine-tune existing MLLMs (\textit{e.g.}, InternVL2.5-8B~\cite{chen2024internvl}). Notably, we find that the synthesized data can significantly boost existing MLLMs (\textit{e.g.}, InternVL2.5-8B +25.0\% on MCTS-VCB and +16.3\% on DREAM-1K~\cite{DBLP:journals/corr/abs-2407-00634}), as shown in Figure~\ref{fig:sft_performance_radio}. 

In summary, our contributions are as follows:
\begin{enumerate}
    \item We propose an automated captioning framework, named AutoCaption, that leverages Monte Carlo Tree Search (MCTS) to iteratively analyze and describe video details, achieving a comprehensive understanding of the video content.
    \item We develop a fine-grained video captioning evaluation benchmark MCTS-VCB for MLLMs. To the best of our knowledge, it is the first MCTS based benchmark in this field. 
    \item We prove that AutoCaption can be applied to generate data for fine-tuning existing MLLMs. Experimental results indicate that it is a more efficient method to boost the captioning capability of MLLMs.
\end{enumerate}

\begin{figure*}[!h]
    \centering
    \includegraphics[width=\linewidth]{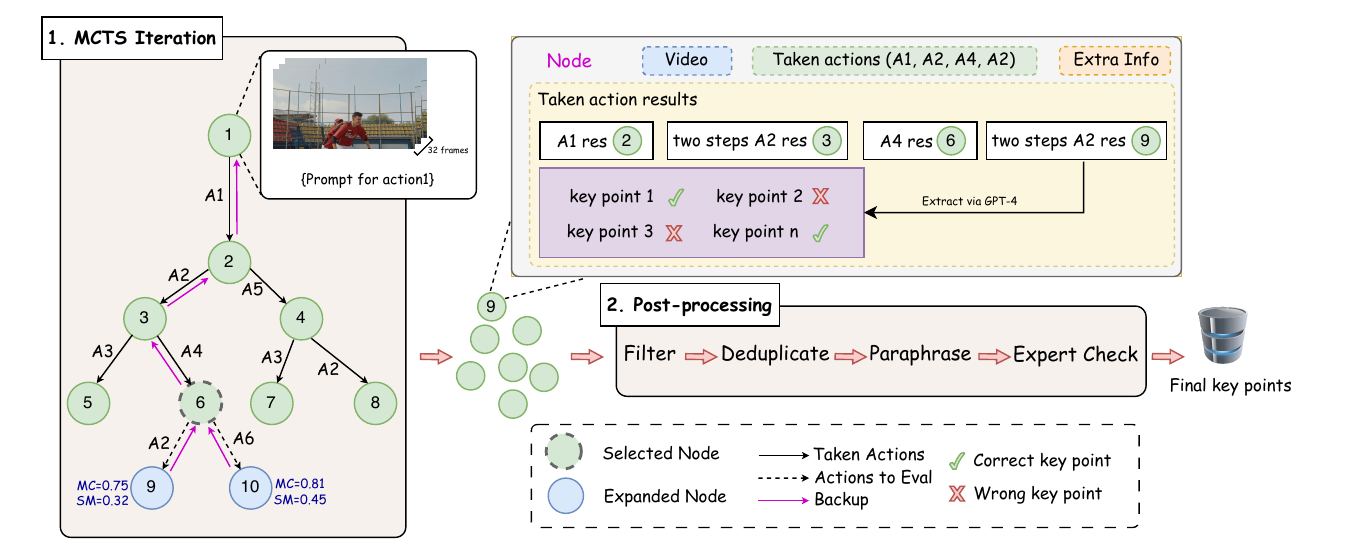}
    \caption{The overall workflow of AutoCaption, illustrating the key points generation process from a video. The ``A1''-``A6'' labels on the edges of search tree \(T\) denote six different actions. The MC value and SM value besides two new nodes expanded from the selected node 6 are calculated in the ``Evaluation'' phase.}
    \label{fig: workflow}
\end{figure*}

\section{Related Work}
\label{sec:related_work}

\paragraph{Data Synthesis and Augmentation}
Synthetic data can effectively alleviate the substantial demand for large-scale training data for LLM \citep{DBLP:journals/corr/abs-2401-15422}. There are two commonly used approaches improving the capabilities of LLMs: data augmentation and data synthesis \citep{wang2024surveydatasynthesisaugmentation}. The former involves further processing of existing data, including data labeling \citep{DBLP:journals/corr/abs-2307-02179, DBLP:journals/corr/abs-2304-06588, DBLP:journals/corr/abs-2304-10145}, data 
transformation~\citep{DBLP:journals/corr/abs-2302-13007, DBLP:conf/nips/DunlapUZYGD23, DBLP:conf/acl/ChenGBS023, shi2024criskevalchinesemultilevelrisk, DBLP:conf/acl/YuL0WLJZSCCLX24}, to create fine-grained data or augmented data. 
The latter typically curates data from scratch with common methods, including model distillation~\citep{DBLP:conf/iclr/XuSZG0FTLJ24, DBLP:journals/corr/abs-2405-14333} and self-improvement~\citep{DBLP:conf/acl/WangKMLSKH23, DBLP:conf/icml/ChenDYJG24}. 
Self-improvement usually involves multiple refinements on sampled data to obtain high-quality synthetic data. Theoretically, tree-search methods like MCTS~\citep{DBLP:conf/nips/YaoYZS00N23, DBLP:conf/emnlp/HaoGMHWWH23} can sample and iterate over an infinite number of paths, that can not only ensure high-quality data but also significantly increase data diversity. 
Owing to this reason, MCTS has been widely applied to complex reasoning tasks \citep{DBLP:journals/corr/abs-2405-03553, DBLP:journals/corr/abs-2406-07394, DBLP:journals/corr/abs-2408-06195, DBLP:journals/corr/abs-2410-02884, DBLP:journals/corr/abs-2406-06592, DBLP:conf/acl/WangLSXDLCWS24} and planning \citep{li2025chatsopsopguidedmctsplanning}. Inspired by its advantage in reasoning, we leverage MCTS to continuously capture the details of videos, ensuring the maximum coverage of video content.

\paragraph{Video Understanding Benchmarks}
Video understanding benchmarks can be broadly categorized into two groups.  For the former, there are benchmarks that evaluate different video understanding tasks, such as spatial-temporal reasoning~\cite{DBLP:conf/aaai/YuXYYZZT19, DBLP:conf/cvpr/JangSYKK17, DBLP:conf/cvpr/ZhangZZWLG20, DBLP:journals/tcsv/TangLLLJJYX22}, causal reasoning~\citep{DBLP:conf/cvpr/XiaoSYC21}, and long video reasoning~\citep{DBLP:conf/cvpr/SongCWZZWCG0ZLH24, DBLP:journals/corr/abs-2404-17176}. For the latter, there are comprehensive evaluation frameworks like MMBench \citep{DBLP:conf/eccv/LiuDZLZZYWHLCL24} and MVBench \citep{DBLP:conf/cvpr/0002WH00LWX0L0024}, which focus on different video understanding tasks by conducting question answering (QA) on various videos. However, these comprehensive frameworks usually struggle to provide a comprehensive evaluation over video captioning.

\paragraph{Video Captioning Benchmarks} As a fundamental task of video understanding, it has been drawing attention to building video caption benchmarks. Initially, these benchmarks are specific to certain domains, such as social media \citep{DBLP:conf/emnlp/GellaLR18}, movies \citep{DBLP:conf/cvpr/RohrbachRTS15, DBLP:journals/ijcv/RohrbachTRTPLCS17}, and cooking \citep{DBLP:journals/tacl/RegneriRWTSP13, DBLP:conf/aaai/ZhouXC18}. There are also open-domain benchmarks, such as MSR-VTT \citep{DBLP:conf/cvpr/XuMYR16} and MSCV \citep{DBLP:conf/acl/ChenD11}, which annotate a video with multiple similar single-sentence descriptions. Vatex~\citep{DBLP:conf/iccv/WangWCLWW19} is similar to these benchmarks but provides bilingual single-sentence descriptions with both Chinese and English. However, such descriptions cannot fully, accurately, and comprehensively convey the content of a video. The emergence of DREAM-1K~\citep{DBLP:journals/corr/abs-2407-00634} provides a more comprehensive evaluation of video captioning through a fine-grained approach, but the key points in a video are manually annotated as ``events'', with an average of only 6.3 key points per video, which may miss details of videos. Our proposed AutoCaption method, can automatically annotate 122.3 key points per video, thus providing a more comprehensive evaluation of the video captioning capability of MLLMs.

\section{Preliminary}
\label{sec:preliminary}

MCTS is a heuristic search algorithm for some kinds of decision processes, such as game AI, planning, and optimization problems. MCTS evaluates the potential value of different decisions by simulating multiple possible future paths in a decision tree. The core idea is to estimate the value of each node through random simulations and gradually expand and optimize the search tree. 

There are four main steps in MCTS~\citep{DBLP:journals/tciaig/BrownePWLCRTPSC12}:
1) \textbf{Selection} Starting from the root node, child nodes are selected based on a certain strategy (\textit{e.g.}, UCB~\cite{DBLP:journals/ior/ChangFHM05}) until a leaf node is reached.
2) \textbf{Expansion}
If the selected node is not in a terminal state, expand it by adding one or more child nodes to the selected node.
3) \textbf{Simulation}
From the newly expanded node, a random simulation is performed until the end of the game or a predefined depth is reached. The result of the simulation is used to estimate the value of the node.
4) \textbf{Backpropagation}
The simulation result is propagated back to all nodes along the path, while the visit count and value estimate of these nodes are updated.

AutoCaption is an automated framework aimed at generating detailed key points that fully cover the content of videos using MCTS. Utilizing the tree search feature of MCTS enables the progressive enhancement of video details such as appearance and comprehensive descriptions of the environment by taking different actions iteratively. Inspired by \citet{DBLP:journals/corr/abs-2408-06195}, we can use a small MLLM as the generator to dig detailed information in our proposed AutoCaption framework. 


\section{AutoCaption}
\label{sec:autocaption}
We divide AutoCaption into two essential procedures: MCTS iteration and data post-processing, as shown in Figure~\ref{fig: workflow}.  We provide the pseudocode of AutoCaption in Algorithm \ref{alg:autocaption} in appendix. In this section, we explain these two procedures in detail. 

Given a video \(v\) and a small MLLM for generating video captions, we can construct a  search tree \(T\) through MCTS. As shown in Figure \ref{fig: workflow}, we assign the video \(v\) to the root node, each edge signifies an executing action \(a\), and a child node \(s\) embodies a state resulting from \(a\). The trajectory \(t(s)\) from the root node to any child node \(s\) symbolizes the process of increasingly concentrating on imperceptible video content to produce new detailed descriptions.


\subsection{MCTS Iteration}
\label{subsec:mcts_iteration}

In the MCTS iteration, each root node contains a video \(v\) that needs to be described, a list of actions taken from the root node to the current node \(s\), and a list of action results (video descriptions) generated by the corresponding actions. Additionally, each node \(s\) contains statistical information: \({N}(s)\) denotes the number of times that the node is visited, \(\text{MC}(s)\) is the Monte Carlo value of the node, which measures the correctness of the action result obtained after executing the corresponding action at node \(s\), and \(\text{SM}(s)\) is the similarity score of the node to all other nodes in trajectory \(t(s)\). Then, we follow the standard MCTS steps to conduct the iteration procedure.


\paragraph{Selection}
In the \(i\)-th iteration of MCTS, we first select a node \(s\) with the highest value (calculated in ``Evaluation''  phase) from all the leaf nodes \(L(T)\) 
in the search tree \(T\) based on a variant of the PUCT algorithm \citep{DBLP:journals/amai/Rosin11} formulated as:
\begin{equation}
    {s}_{i}=\arg \max _{\mathbf{s} \in L(T)}\left[{Q}\left({s}, {a}\right)+c \frac{\sqrt{N_{\text {parent }}(s)}}{1+N\left({s}\right)}\right]
\end{equation}
where \({Q}\left({s}, {a}\right)\) is the state value function after taking action \(a\),
with a higher value indicating that the selected node is more likely to generate a correct detailed description when executing corresponding actions. 
Refer to \citet{DBLP:journals/corr/abs-2406-06592}, $c$ is a constant set to 0.125 that controls the exploration degree, \(N_{\text{parent}}(s)\) denotes the current number of visit times to the parent node of \(s\). 

\paragraph{Expansion}
Unlike tranditional MCTS, we merge ``simulation'' phase into the ``Expansaion'' phase since excuating an action by MMLM can be treated as ``simulation''.
In the \(i\)-th iteration, we expand the possible actions at node \(s_i\) selected in the previous phase. Specifically, we have defined six types of actions: \textit{Overall Description} (A1), \textit{Detail Description} (A2), \textit{Temporal Perspective Description} (A3), \textit{Spatial Perspective Description} (A4), \textit{Background Description} (A5), and \textit{Camera Movement Description} (A6). These actions follow a certain execution order. The \textit{Overall Description} action is executed only once and only after the root node. To give MCTS iteration a good initialization, the \textit{Overall Description} action is executed by GPT-4o~\cite{openai2024gpt4o} and Gemini-1.5-Pro~\cite{gemini2024gemini15}. These six actions comprehensively cover the description of video content from various perspectives. Meanwhile, these actions can ensure the breadth of descriptions while progressively increasing the depth of detail through continuous iteration. 
In the \(i\)-th iteration, for the selected child node \(s_i\), since we expect the generator to focus on different details when executing the corresponding action \(a_j\), we always include the correct key points extracted from the results of previously executed actions in \(a_j\) in trajectory \(t(s_i)\) into the action prompt to enforce the generator paying attention to new details. 
Therefore, each node \(s\) in \(t(s)\) needs to store both the historical actions and action results. 

For the selected node \(s_i\), we do not expand all actions but randomly expand only two of them under the condition that the probability of sampling the \textit{Detailed Description} action is twice that of sampling other actions. This is because, for video captioning, it is not necessary to generate multiple paths to achieve a higher completion rate as in reasoning tasks. Instead, video captioning requires a more comprehensive description of video details during executing \textit{Detail Description} action. Additionally, for the rest of the actions, we observe that the improvement in detail brought by multiple samplings is very limited, so it is unnecessary to expand all actions at each node \(s_i\). 

In addition, to enable the generator to continuously discover new details, we have adopted a two-stage process for the \textit{Detail Description} action: 1) First, prompting the generator to identify a new detail besides those previously found. Assuming response in this stage to be \(\text{Ans}_{s1} \). 2) Then, we utilize Qwen2.5-7B-Instruct~\cite{qwen2_5} to extract the details that need further observation from \(\text{Ans}_{s1} \), along with the category of the detail and its specific attributes, integrating them into a prompt to guide the generator to continue describing new details. The two-stage process effectively reduces repetitive descriptions. The prompts corresponding to different actions are shown in Appendix~\ref{appendix: Prompts Utilized for Different Action Types}.

\paragraph{Evaluation}
\label{sec: evaluation}

\begin{figure}[!t]
    \centering
    \includegraphics[width=\linewidth]{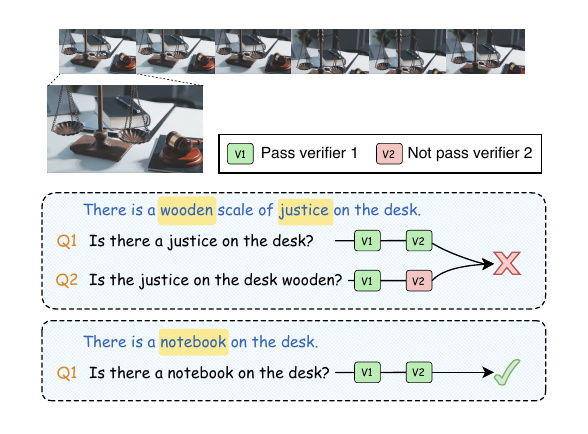}
    \caption{Illustration of the key point verification process. Texts in yellow shade are key information need to be verified. Each key information needs a verification question. ``V1'' and ``V2'' denote GPT-4o and Qwen2-VL-72B, respectively.}
    \label{fig: kp_verify}
\end{figure}

We evaluate all the new nodes expanded during the ``Expansion'' phase here. 
The state value of each node is calculated using the following formula adopted in \citet{DBLP:journals/corr/abs-2406-06592}:
\begin{equation}
    Q(s, a)=\alpha^{1-\mathrm{MC}(s)} \cdot \beta^{\mathrm{SM}(s)}
\end{equation}
where \(\alpha\) and \(\beta\) are values \(\in\) \((0,1)\), used to balance the value and similarity. In \citet{DBLP:journals/corr/abs-2406-06592}, \(\alpha=0.5\) and \(\beta=0.9\), while both \(\alpha\) and \(\beta\) are set to 0.5 in our experiment since \(\mathrm{MC}(s)\) and \(\mathrm{SM}(s)\) are same important. Here, \(\mathrm{MC}(s)\) represents the Monte Carlo simulation value of the node \(s\), and \(\mathrm{SM}(s)\) represents the similarity of the node \(s\). The node with higher \(\mathrm{MC}(s)\) and lower \(\mathrm{SM}(s)\) will have a higher state value \(Q(s, a)\). The calculations for \(\mathrm{MC}(s)\) and \(\mathrm{SM}(s)\) are as follows:
\begin{equation}
    \text{SM}(s) = \frac{1}{|t(s)|-1}\sum_{\substack{u \in t(s) \\ u \neq s}}\text{sim}(u, s) 
\end{equation}
\begin{equation}
    \text {MC}(s)=\frac{N_{\text{correct}}(s)} {N_{\text{total}}(s)}
\end{equation}
where \(N_{\text{correct}}(s)\) is the number of key points passed the verification and \(N_{\text{total}}(s)\) is the total number of key points extracted from node \(s\).

\begin{table*}[!t]
    \centering
    \scriptsize
    \begin{tabularx}{\linewidth}{ 
			>{\raggedright\arraybackslash\hsize=1.35\hsize\linewidth=\hsize}X
			>{\centering\arraybackslash\hsize=0.92\hsize\linewidth=\hsize}X
			>{\centering\arraybackslash\hsize=0.92\hsize\linewidth=\hsize}X
			>{\centering\arraybackslash\hsize=0.92\hsize\linewidth=\hsize}X
			>{\centering\arraybackslash\hsize=0.92\hsize\linewidth=\hsize}X
			>{\centering\arraybackslash\hsize=1.05\hsize\linewidth=\hsize}X
			>{\centering\arraybackslash\hsize=0.92\hsize\linewidth=\hsize}X
		}
        \toprule
        {Model Name} & \makecell{Appearance \\ Description} & \makecell{Action \\ Description} & \makecell{Environment \\ Description} & \makecell{Object \\ Description} & \makecell{Camera Movement \\ and Composition} & {Overall} \\
        \midrule
\rowcolor{tabelcellgrey} \multicolumn{7}{l}{\textbf{\textit{Open-Source MLLMs}}} \\
InternVL2\_5-8B & 65.1/36.5/46.8 &  65.0/42.0/51.0 &  76.0/48.7/59.4 &  72.1/30.3/42.7 &  54.7/31.7/40.1 &  69.6/40.1/50.8   \\
InternVL2\_5-26B & 65.2/37.1/47.3 &  67.6/43.4/52.8 &  75.6/49.1/59.5 &  73.6/31.5/44.1 &  54.4/32.0/40.3 &  70.1/40.9/51.7   \\
InternVL2\_5-38B & 65.0/37.0/47.1 &  68.2/44.0/53.5 &  75.7/48.9/59.5 &  73.8/31.7/44.4 &  56.2/31.7/40.6 &  70.6/41.0/51.8   \\
InternVL2\_5-78B & 65.3/38.2/48.2 &  67.7/44.7/53.8 &  76.4/49.8/60.3 &  73.5/31.8/44.4 &  55.6/32.0/40.6 &  70.6/41.6/52.4   \\
LLaVa\_OV\_Qwen2\_7B & 75.8/45.1/56.5 &  77.3/58.9/66.8 &  84.5/59.1/69.6 &  77.1/42.2/54.5 &  73.2/42.4/53.7 &  79.6/51.8/62.8   \\
LLaVa\_OV\_Qwen2\_72B & 75.8/45.5/56.9 &  78.9/\underline{60.3}/\underline{68.3} &  \textbf{86.1}/60.2/\underline{70.9} &  78.7/\underline{43.1}/\underline{55.7} &  \underline{79.6}/\underline{45.3}/\underline{57.7} &  \underline{81.3}/\underline{52.9}/\underline{64.1}   \\
LLaVa\_Video\_Qwen2\_7B & 78.0/41.3/54.0 &  79.7/50.7/61.9 &  82.7/58.4/68.5 &  76.4/41.0/53.4 &  76.1/40.8/53.1 &  79.7/48.9/60.6   \\
LLaVa\_Video\_Qwen2\_72B & \textbf{79.8}/\underline{45.8}/\underline{58.2} &  \underline{80.9}/57.9/67.5 &  85.2/59.1/69.8 &  \underline{80.2}/41.5/54.7 &  77.9/44.1/56.3 &  \textbf{82.1}/51.7/63.4   \\
MiniCPM-V-2\_6 & 75.3/40.8/52.9 &  73.0/57.8/64.5 &  82.7/58.2/68.3 &  78.3/38.9/52.0 &  75.6/\textbf{53.9}/\textbf{62.9} &  77.7/50.4/61.2   \\
PLLaVA-7B & 47.4/28.3/35.5 &  47.6/37.6/42.0 &  69.0/37.7/48.7 &  61.9/17.6/27.4 &  37.6/20.3/26.4 &  55.1/30.4/39.2   \\
PLLaVA-13B & 69.9/35.9/47.4 &  69.5/44.0/53.9 &  78.0/46.2/58.0 &  66.7/27.3/38.8 &  69.8/29.5/41.4 &  72.5/38.7/50.4   \\
PLLaVA-34B & 72.3/33.5/45.8 &  73.2/44.6/55.4 &  78.3/46.6/58.4 &  73.6/27.9/40.4 &  65.6/26.5/37.7 &  74.9/38.5/50.9   \\
Qwen2-VL-7B & 74.4/43.5/54.9 &  75.8/57.7/65.5 &  80.4/60.2/68.8 &  77.6/40.7/53.4 &  71.8/34.9/46.9 &  77.8/50.9/61.5   \\
Qwen2-VL-72B & 71.4/45.8/55.8 &  75.9/57.9/65.7 &  80.9/\underline{60.8}/69.4 &  78.7/42.5/55.2 &  73.2/37.7/49.7 &  77.7/52.1/62.4   \\
Tarsier-7B & 76.3/34.0/47.1 &  69.7/49.7/58.0 &  \underline{85.7}/56.4/68.0 &  72.9/35.5/47.8 &  58.7/36.0/44.6 &  76.7/45.2/56.9   \\
Tarsier-34B & \textbf{79.8}/34.3/48.0 &  75.5/51.5/61.2 &  \textbf{\underline{86.8}}/55.9/68.0 &  76.8/36.4/49.4 &  62.1/35.8/45.5 &  79.6/45.7/58.0   \\
Llama-3-VILA1.5-8B & 71.7/39.2/50.7 &  73.8/46.6/57.1 &  80.9/52.3/63.5 &  73.5/37.1/49.3 &  66.9/31.6/42.9 &  75.6/44.2/55.8   \\
VILA1.5-13B & 69.4/39.0/50.0 &  70.5/50.3/58.7 &  79.3/51.6/62.5 &  74.0/34.6/47.2 &  66.8/35.2/46.1 &  73.9/44.4/55.4   \\
VILA1.5-40B & 74.5/42.3/54.0 &  76.1/55.8/64.4 &  82.1/53.4/64.7 &  79.0/37.9/51.2 &  73.1/39.8/51.6 &  78.3/47.6/59.2   \\
\rowcolor{tabelcellgrey} \multicolumn{7}{l}{\textbf{\textit{Closed-Source MLLMs}}} \\
Gemini-1.5-Pro & \underline{78.2}/\textbf{\underline{59.9}}/\textbf{\underline{67.9}} &  \textbf{82.4}/\textbf{\underline{64.8}}/\textbf{72.5} &  81.0/\textbf{\underline{70.7}}/\textbf{\underline{75.5}} &  \textbf{82.0}/\textbf{\underline{56.0}}/\textbf{\underline{66.5}} &  \textbf{79.9}/\textbf{\underline{59.0}}/\textbf{\underline{67.9}} &  80.9/\textbf{\underline{63.6}}/\textbf{\underline{71.2}}   \\
GPT-4o & \textbf{\underline{84.9}}/\textbf{47.2}/\textbf{60.6} &  \textbf{\underline{86.4}}/\textbf{62.7}/\textbf{\underline{72.7}} &  80.9/\textbf{67.1}/\textbf{73.4} &  \textbf{\underline{87.3}}/\textbf{50.6}/\textbf{64.1} &  \textbf{\underline{81.9}}/40.8/54.5 &  \textbf{\underline{83.8}}/\textbf{57.5}/\textbf{68.2}   \\
        \bottomrule
    \end{tabularx}
    \caption{Performance on MCTS-VCB across different key point categories. Each cell contains ``\textbf{Precision} / \textbf{Recall} / \textbf{F1 Score}\textquotedblright. The best, second-best, third best results are 
 \textbf{\underline{bold underlined}}, \textbf{bold} and \underline{underlined} respectively.}
    \label{tab: performance across kp categories}
\end{table*}

The key points are fine-grained key information extracted from a node \(s\) by Qwen2-VL-72B~\cite{qwen2_5-VL} under the prompt displayed in Figure~\ref{fig: prompt-extract kp} in the Appendix, ensuring that each key point contains only atomic-level subjects and indivisible predicates. 
We further categorize each extracted atomic-level key point into five categories: \textit{Appearance Description}, \textit{Action Description}, \textit{Environment Description}, \textit{Object Description}, and \textit{Camera Movement and Composition}. We provide an example video with these five key point types in Figure~\ref{fig: MCTS-VCD-kp_category} in the Appendix.

To verify each key point, we first determine the key information that needs to be verified, such as actions, appearance, attributes of an object, etc. Then, we utilize GPT-4~\cite{openai2024gpt4technicalreport} to construct a ``yes/no'' question for each key information needed to be verified and then use two verifiers (\textit{i.e.}, GPT-4o and Qwen2-VL-72B) to answer them. If all yes/no questions derived from a key point are shown to be ``yes'' for both verifiers, we consider the key point passing the verification. Figure~\ref{fig: kp_verify} shows the verification process of a key point. Some more explored
alternative verification methods, which are not effective, are provided in Appendix \ref{appendix: Evaluation Details}.


We claim that the \(\mathrm{SM}(s)\) in \(Q(s, a)\) is necessary because node \(s\) with high similarity to nodes in \(t(s)\) is more likely to produce repetitive results. This not only undermines data diversity but also increases unnecessary validation.

\paragraph{Backpropagation}
At the end of the \(i\)-th iteration, starting from the selected node \(s_i\), we update the visit times \(N(s)\) and state value \(Q(s, a)\) of all nodes in \(t(s_i)\) except node \(s_i\) according to the following backpropagation rules:
\begin{equation}
\begin{aligned}
    N(s) & \leftarrow N(s) + 1 \\
    Q(s, a) & \leftarrow \frac{1}{N(s)}\sum_{u \in s.\text{children}} Q(u, a_u)
\end{aligned}
\end{equation}
where \(a_u\) is the action that leads to node \(u\).

\subsection{Data Post-processing}
\label{subsec:data_post_process}
\label{section: Data Post Processing}
After the MCTS iterations over a video, we construct a search tree \(T\) for the video. We then obtain all nodes in \(T\), where each node, except the root node, contains a description generated by the corresponding actions. These descriptions need to be post-processed (\textit{e.g.}, filtered, deduplicated, paraphrased and human checked) to form final key points of the video. Note that the post-processing is lightweight that requires low computation resources.The post-processing detail is described in Appendix \ref{appendix: Post Processing Details}.

\section{MCTS-VCB}
\label{sec:mcts_vcb_benchmark}
MCTS-VCB is a video caption benchmark curated with AutoCaption described in Section~\ref{sec:autocaption}. We use Qwen2-VL-7B \citep{DBLP:journals/corr/abs-2409-12191} as the generator in MCTS iteration. We uniformly sample 64 frames from each video, and the iteration number is set to 25 since we found that more duplicated contents are produced with more iteration.

\paragraph{Data Source}
Initially, we randomly select 2,000 video clips from open-soure data to form the video candidates for MCTS-VCB. Among these videos, we extract visual and textual information to automatically exclude clips lacking distinct actions or subjects. The majority of excluded clips feature extensive special effects, an absence of clear subjects, or indeterminate objectives. The filtering process is detailed in Appendix \ref{appendix: Video Screening Detail}.

\begin{figure*}[!ht]
    \centering
    \includegraphics[width=\linewidth]{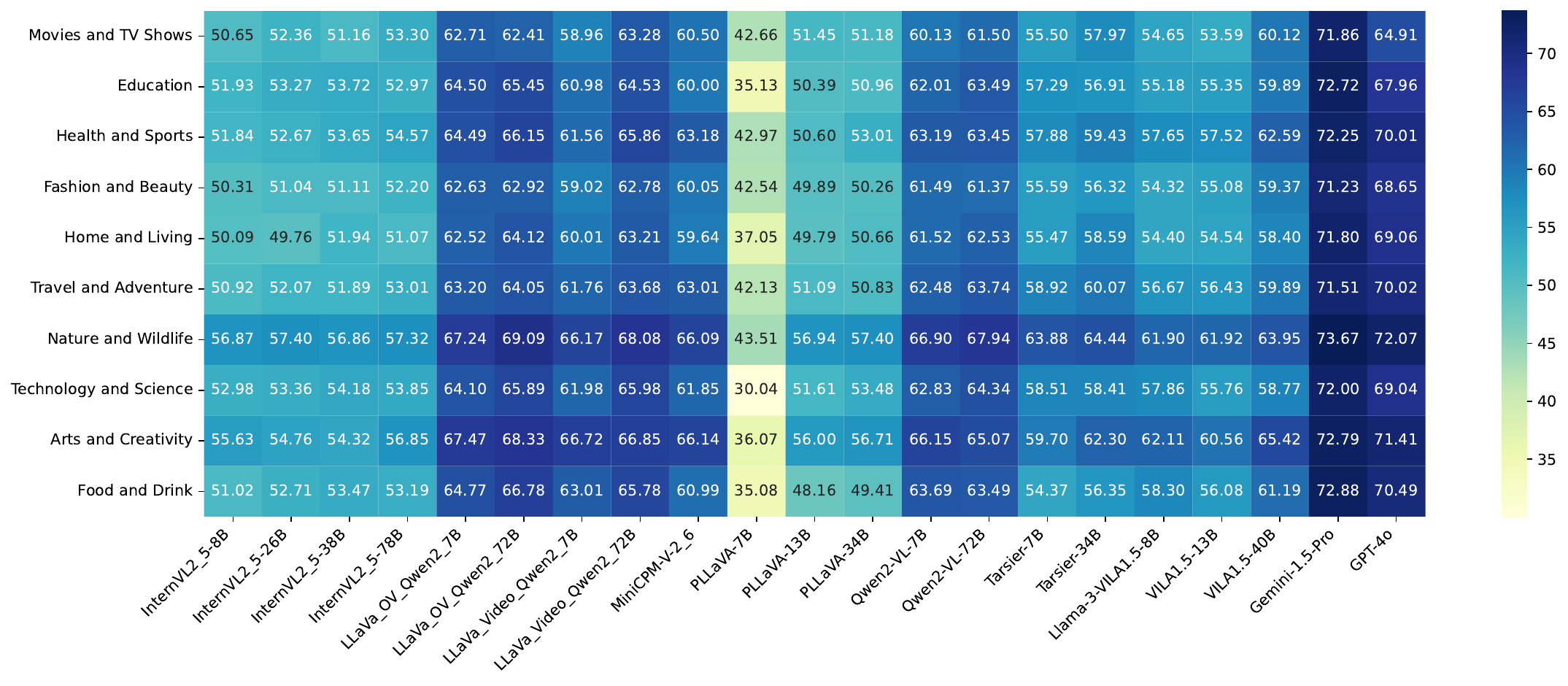}
    \caption{F1 score across different video categories.}
    \label{fig: f1_score_across_video_categories}
\end{figure*}

\paragraph{MCTS-VCB Statistics}
The average duration of these clips is 12.39 seconds, which ensures a balanced amount of information is presented in each, making them suitable for evaluating captioning tasks. The distribution of video durations is shown in Figure~\ref{fig:video_length_bar} in Appendix~\ref{appendix: MCTS-VCB Statistics}. MCTS-VCB encompasses 10 categories of videos, 5 categories of video content key points, enabling a thorough evaluation of MLLMs' capability to describe various video genres. The detailed explanation and statistics of different video categories and key point categories are illustrated in Appendix \ref{appendix: MCTS-VCB Statistics}.

\section{Experiments}
\label{sec:experiments}
We evaluated 21 open- and closed-source MLLMs, with sizes ranging from 7B to 78B. We conducted a comprehensive evaluation of these MLLMs across different dimensions. The model cards of the 21 evaluated MLLMs and corresponding experimental settings (such as the number of frames and used prompts) can be found in Appendix~\ref{appendix: experiment_setting}. 
To further validate the effectiveness of AutoCaption, we conducted a series of augmented experiments. 

\paragraph{Evaluation Metrics}
We used settings similar to those adopted in the evaluation method proposed by DREAM-1K \citep{DBLP:journals/corr/abs-2407-00634}. First, for each evaluated model, we used the prompt shown in Figure \ref{fig: prompt-extract kp} to guide Qwen2-VL-72B in extracting key points \(KP_{\text{model}}\) from the model-generated description \(D_{\text{model}}\). For each video, MCTS-VCB contains the verified video key points \(KP_{\text{ref}}\).
Then, we calculated the ratio of key points in \(KP_{\text{model}}\) that are entailed in \(KP_{\text{ref}}\) as \textbf{Precision}, and the ratio of key points in \(KP_{\text{ref}}\) that are entailed in \(D_{\text{model}}\) as \textbf{Recall}. Finally, we calculated \textbf{F1} by \( \frac{2*P*R}{P+R}\). This evaluation method allows us to assess both the accuracy and completeness of the responses from all evaluated MLLMs. We used Qwen2.5-72B-Instruct to be the judge model and the prompt is displayed in Figure \ref{fig: prompt-determine_entail_contra_netural}.

\subsection{Evaluation Results}
\label{subsec:evaluation_results}

\textbf{Performance across different key point types.} 
Table \ref{tab: performance across kp categories} presents the performance of different MLLMs on MCTS-VCB. Overall, Gemini-1.5-Pro excels in multiple categories, achieving 4 wins out of 5 categories and overall best performance, demonstrating its robust capability in handling video caption tasks. GPT-4o follows closely, performing well in the action description category, highlighting its strengths in these specific tasks. Though the gap between open- and closed-source MLLMs still exists, LLaVa\_OV\_Qwen2\_72B performs best among the open-source MLLMs with an overall F1 score of 64.1. Furthermore, we can observe that performance generally improves along with increasing the model size. Surprisingly, the InternVL series MLLMs demonstrate merely average performance; this will be further investigated Section~\ref{section: Augmented Experiments}.

\paragraph{Performance across different video categories.}
Figure \ref{fig: f1_score_across_video_categories} shows the performance of various MLLMs across different video categories. It is evident that MLLMs achieve better results in the ``nature and Wildlife'' and ``Arts and Creativity'' categories, possibly due to these videos' real-life resemblance, facilitating model learning. In contrast, performance in the remaining categories varies slightly. Although videos in ``Movie and TV Shows'' offer easily accessible training data with clear subjects and rich content, these types of videos often contain more detailed information. In our MCTS-VCB benchmark, these detailed information is captured thoroughly. However, existing MLLMs struggle to process the detailed information presented in these videos, leading to less impressive performance. 

\paragraph{Performance on different frame numbers.}
As shown in Figure~\ref{tab:perform_frames}, we evaluated four MLLMs with a similar model size to investigate the performance when varying numbers of input frames. During the training process, LLaVa\_OV\_Qwen2\_7B is trained with a maximum frame number of 32 while LLaVa\_Video\_Qwen2\_7B and InternVL2\_5-8B are trained with a maximum frame number of 64. Consequently, we observe that the performance of these three models increases as the number of frames increases, until the frame number exceeds the maximum frame number used during training. MiniCPM-V-2\_6 exhibits the strongest performance at 16 frames, which we attribute to its lack of training on sufficiently long contexts. This suggests that, as more input frames are provided, models can acquire more information and detect finer details, but this requires prior training on long context length.

\paragraph{Performance across different similarity threshold.}
\label{section: Performance across different similarity threshold}
We explored the impact of using different similarity thresholds for merging data from each node on the evaluation results. 
We conducted experiments by uniformly sampling 11 similarity thresholds in [0.7, 0.9]. 
The results indicate that as the similarity threshold increases, the MLLMs' performance also improves. However, this is an expected phenomenon, and the relative ranking remains unchanged, demonstrating the accuracy of the evaluation metrics.
Results of this group of experiments are provided in Appendix \ref{appendix: Supplementary Experiment Results}.

\paragraph{Human Consistency}
We selected 100 videos from the 1,765 videos in MCTS-VCB and used 3 MLLMs to check their consistency with human evaluations across various methods. For each MLLM and video, annotators reviewed key points from AutoCaption and the MLLM's output to assess relationship between these two sets of key points (entailment, neutral, contradiction) and calculated the F1 score.

We employed four different evaluation metrics, two of which are based on n-gram overlap (Bert-Score \cite{DBLP:conf/iclr/ZhangKWWA20} and ROUGE \citep{lin-2004-rouge}), and the other two are based on judgment models (CLAIR \citep{DBLP:journals/corr/abs-2310-12971} and AutoDQ \citep{DBLP:journals/corr/abs-2407-00634}). We calculated Kendall's $\tau$, Spearman's $\rho$, and Pearson $r$ correlation scores. Below are the consistencies of the selected 3 models with human evaluation across the four assessment methods. All P-values < 0.05.  As shown in Table \ref{tab:grouped-model-results}, our evaluation method ensures the highest consistency, demonstrating MCTS-VCB's superiority.

\begin{table}[!t]
\centering
\scriptsize
\begin{tabular}{lccc}
\toprule
\multicolumn{4}{l}{\textbf{InternVL2\_5-8B}} \\
\midrule
\textbf{Method} & Kendall's $\tau$ & Spearman's $\rho$ & Pearson $r$ \\
Bert-Score      & 18.8             & 27.9              & 27.1 \\
ROUGE           & 17.3             & 23.5              & 24.7 \\
CLAIR           & 41.8             & 53.2              & 58.2 \\
AutoDQ          & 32.4             & 47.4              & 43.6 \\
\textbf{MCTS-VCB (ours)} & \textbf{68.4} & \textbf{87.6} & \textbf{88.5} \\
\midrule
\multicolumn{4}{l}{\textbf{LLaVa\_OV\_Qwen2\_72B}} \\
\midrule
\textbf{Method} & Kendall's $\tau$ & Spearman's $\rho$ & Pearson $r$ \\
Bert-Score      & 18.9             & 28.5              & 31.3 \\
ROUGE           & 20.5             & 30.3              & 33.6 \\
CLAIR           & 25.6             & 32.1              & 39.2 \\
AutoDQ          & 20.8             & 28.7              & 24.5 \\
\textbf{MCTS-VCB (ours)} & \textbf{56.8} & \textbf{75.0} & \textbf{80.5} \\
\midrule
\multicolumn{4}{l}{\textbf{PLLaVA-7B}} \\
\midrule
\textbf{Method} & Kendall's $\tau$ & Spearman's $\rho$ & Pearson $r$ \\
Bert-Score      & 36.3             & 52.0              & 61.8 \\
ROUGE           & 36.2             & 51.8              & 57.7 \\
CLAIR           & 53.0             & 67.7              & 70.4 \\
AutoDQ          & 33.2             & 45.1              & 48.1 \\
\textbf{MCTS-VCB (ours)} & \textbf{77.8} & \textbf{92.1} & \textbf{94.8} \\
\bottomrule
\end{tabular}
\caption{Correlation results (Kendall’s $\tau$, Spearman’s $\rho$, and Pearson $r$) of various evaluation metrics on three models.}
\label{tab:grouped-model-results}
\end{table}

\begin{table}[!t]
    \centering
    \scriptsize
    \begin{tabularx}{\linewidth}{ 
        >{\raggedright\arraybackslash\hsize=2.2\hsize\linewidth=\hsize}X
        >{\centering\arraybackslash\hsize=0.7\hsize\linewidth=\hsize}X
        >{\centering\arraybackslash\hsize=0.7\hsize\linewidth=\hsize}X
        >{\centering\arraybackslash\hsize=0.7\hsize\linewidth=\hsize}X
        >{\centering\arraybackslash\hsize=0.7\hsize\linewidth=\hsize}X
    }
        \toprule
        \multicolumn{1}{c}{\multirow{2}{*}{Models}} & \multicolumn{4}{c}{\#Frames} \\
        \cmidrule(l){2-5}
        & 8 & 16 & 32 & 64 \\
        \midrule
        LLaVa\_OV\_Qwen2\_7B & 62.1 & 62.8 & 62.9 & 62.6 \\
        LLaVa\_Video\_Qwen2\_7B & 59.2 & 60.6 & 61.7 & 63.5 \\
        MiniCPM-V-2\_6 & 60.3 & 61.2 & 60.2 & 58.8 \\
        InternVL2\_5-8B & 50.4 & 50.8 & 51.2 & 51.5 \\
        \bottomrule
    \end{tabularx}
    \caption{Performance on MCTS-VCB across different
input frame numbers.}
    \label{tab:perform_frames}
\end{table}

\subsection{Ablation Study}
\label{section: Augmented Experiments}
\textbf{AutoCaption v.s. Human Annotation}
We extracted 50 videos and their corresponding human-annotated results (\(KP_{\text{human}}\)) from DREAM-1K and obtained the automatic annotations (\(KP_{\text{ac}}\)) for these videos using AutoCaption. We then used GPT-4 to determine the proportion of mutual inclusion between \(KP_{\text{human}}\) and \(KP_{\text{ac}}\). We find that 73.29\% of \(KP_{\text{human}}\) appear in \(KP_{ac}\), while only 14.98\% of \(KP_{\text{ac}}\) appear in \(KP_{\text{human}}\), demonstrating the comprehensiveness of AutoCaption in capturing video content.

\paragraph{MCTS v.s. Beam Search}
In AutoCaption, we use MCTS to iteratively obtain detailed information about the videos. We hence investigated whether using Beam Search could construct key points with the same completeness as the AutoCaption method. We randomly selected 50 videos from MCTS-VCB. We set the beam size to 2, selecting the leaf node with the highest \(Q(s,a)\) for expansion each time, with 25 rounds of expansion. After obtaining these nodes, we perform the data post-processing described in Section \ref{section: Data Post Processing} to obtain the key points (\(KP_{\text{bs}}\)). We again use GPT-4 to determine the proportion of mutual inclusion between \(KP_{\text{bs}}\) and the corresponding key points of the videos in MCTS-VCB (\(KP_{\text{mcts}}\)). We find that 91.62\% of \(KP_{\text{bs}}\) appear in \(KP_{\text{mcts}}\), while only 83.43\% of \(KP_{\text{mcts}}\) appear in \(KP_{\text{bs}}\). This is because Beam Search cannot better expand the breadth of the search. Additionally, since Beam Search only selects the optimal node from the newly expanded nodes for further expansion each time, we find that it has a higher probability of re-executing previously performed actions with identical prompts, leading to redundant information. In contrast, AutoCaption can consider different paths, thereby expanding the search for different details of the video. Furthermore, we find that as the number of Beam Search layers increases, the proportion of repeated content generated by the generator increases, further limiting the acquisition of information.

\paragraph{AutoCaption for Fine-tuning MLLMs}
We use AutoCaption to construct additional training data based on open-source video data. Before synthesizing with AutoCaption, the videos were filtered according to the method described in Appendix \ref{appendix: Video Screening Detail}, resulting in a final set of 9,419 videos.

For each video, key points were generated with AutoCaption. Next, GPT-4 was used to integrate these key points into a detailed thought process, simulating the human process of continuously exploring new details. Subsequently, GPT-4o was used to generate new captions based on the video (uniformly sampled at 32 frames) and the thought process, which were served as fine-tuning data. Detailed construction processes and data examples can be found in Appendix \ref{appendix: Training Data Curation Detail  Example}.

Based on the evaluation results, we selected InternVL2\_5-8B, which had average performance on MCTS-VCB, as the MLLM for fine-tuning. The model was fine-tuned for one epoch. In the experiments, we not only evaluate the results of fine-tuning using data generated by AutoCaption but also compare these results with those obtained by directly fine-tuning on responses generated by GPT-4o. This comparison is made to determine the effectiveness of our method, considering that GPT-4o is used in the process of generating fine-tuning data.
\begin{table}[!t]
    \centering
    \scriptsize
    \begin{tabularx}{\linewidth}{lccc}
        \toprule
        \textbf{Metric} & \textbf{Vanilla} & \textbf{\makecell{+ GPT-4o \\ caption}} & \textbf{\makecell{+AutoCaption \\(Ours)}} \\
        \midrule
        MCTS-VCB Precision & 69.6 & 77.6 & \textbf{78.0} \\
        MCTS-VCB Recall & 40.1 & 52.0 & \textbf{53.6} \\
        MCTS-VCB F1 & 50.8 & 62.3 & \textbf{63.5} \\
        \midrule
        DREAM-1K Precision & 34.4 & 33.7 & \textbf{36.8} \\
        DREAM-1K Recall & 24.7 & 29.0 & \textbf{30.7} \\
        DREAM-1K F1 & 28.8 & 31.2 & \textbf{33.5} \\
        \midrule
        Video-MME & 57.8 & 58.5 & \textbf{59.9} \\
        MMBench-Video & 1.37 & 1.44 & \textbf{1.59} \\
        \bottomrule
    \end{tabularx}
    \caption{Performance improvement of fine-tuned model on out-of-domain and in-domain benchmarks.}
    \label{tab: sft_performance}
\end{table}

Finally, we take MCTS-VCB as an in-domain benchmark and DREAM-1K as the out-of-domain benchmark. Experiment results are shown in Table \ref{tab: sft_performance}. The performance improvement of different dimensions on these two benchmarks is shown in Figure \ref{fig:sft_performance_radio}. Specific data can be found in Appendix \ref{appendix: Complete Experimental Results}. We observe that the training data constructed by our method achieves the highest performance on both benchmarks. According to the official leaderboard~\footnote{\url{https://tarsier-vlm.github.io/}} of DREAM-1K, the fine-tuning results by our methods (\textit{i.e.}, using MCTS and Qwen2-VL-7B as generator) for constructing training data surpassed Qwen2-VL-72B and closely approached GPT-4o-mini. On the MCTS-VCB leaderboard, it ranks second among the open-source MLLMs. Additionally, the fine-tuning results based on data generated by AutoCaption also exceed the results of fine-tuning directly using GPT-4o's responses, validating the effectiveness of AutoCaption.

\section{Conclusion}
We have presented AutoCaption, an automated framework leveraging MCTS to iteratively construct plenty of diverse and descriptive sentences. Additionally, we have constructed a video caption benchmark, MCTS-VCB, with AutoCaption. We evaluate 21 open- and closed-source MLLMs and conduct a comprehensive analysis over various dimensions. Furthermore, we reveal that the MLLM fine-tuned by AutoCaption-generated data achieves a great performance improvement on both in-domain and out-of-domain benchmarks, indicating the effectiveness of AutoCaption.

\section*{Acknowledgments}
The present research was supported by the National Key Research and Development Program of China  (Grant No. 2024YFE0203000). We would like to thank the anonymous reviewers for their insightful comments.

\section*{Limitations}
While AutoCaption constructs a large number of key points, there are inevitably similar key points. Although our analysis shows that different similarity thresholds do not affect the evaluation ranking, it is still necessary to find an appropriate threshold that can balance the resource consumption and accuracy of the evaluation.

\section*{Ethics Policy}
During the process of filtering video clips, we conducted a rigorous review to ensure the safety of the video content and its alignment with human values.

\bibliography{custom}

\appendix

\clearpage

\section{AutoCaption}
\label{appendix: AutoCaption Curation Detail}

\subsection{Video Screening Details}
\label{appendix: Video Screening Detail}
Some video clips, like those lacking a clear subject, featuring prolonged static scenes, or being overly complex and chaotic, are not suitable for captioning tasks. We employ both visual and textual information for filtering purposes. Specifically, we first use GPT-4o to describe all videos in detail, with the prompt ``Please describe the video in detail.''. Subsequently, GPT-4o evaluates each video clip along with its corresponding caption to determine if it is appropriate for captioning, using the prompt as referenced in Figure \ref{fig: prompt-filter video}.

\subsection{Prompts Utilized for Different Action Types}
\label{appendix: Prompts Utilized for Different Action Types}
During the expansion phase, we define six actions, ensuring the extraction of video content details both in terms of breadth and depth. For each action, there is a corresponding prompt to guide the process. The six actions and their respective prompts are described in Table \ref{tab: Action type and prompt}.

\begin{table*}[!t]
    \centering
    \small
    \begin{tabularx}{\linewidth}{ 
			>{\raggedright\arraybackslash\hsize=.3\hsize\linewidth=\hsize}X
			>{\raggedright\arraybackslash\hsize=1.8\hsize\linewidth=\hsize}X
		}  
        \toprule 
       Action Type &  Prompt to Take Corresponding Action\\
        \midrule 
        Overall Description  & Please describe the video in detail. Focus on providing a comprehensive overview of the entire video content, including the main subjects, actions, and settings.  \\
        \midrule 
        Detail Description & \textbf{Stage 1} Next, I think you should carefully observe the details of the page, I think there are these details to consider, People or Animals, Plants, Food, Buildings, Vehicles, Other Objects and so on. Please ouput only one detail you want to focus on. [\textit{Optional} I want you to focus on the {detail} in the video and describe it in the following detail: {relevant\_detail\_attributes}]
        
        \textbf{Stage 2} For MLLMs, generating video content description is a very important task, which is called ``video caption'' task. At the same time, it is hoped that the model can pay more attention to the details of the video when describing the video. \texttt{\textbackslash n} 
        To do that, I'll start by giving the model the video to determine the details that need attention.  \texttt{\textbackslash n} 
        Then I will give you the ``MODEL ANSWER'' (what the model think itself should focus on) and need you to extract the following information from this ``MODEL ANSWER'': \texttt{\textbackslash n}
        1. What video detail does the model need to focus on? \texttt{\textbackslash n}
        2. What category does this detail fall into? \texttt{\textbackslash n}
        3. What describe aspects can model focus on for this detail? \texttt{\textbackslash n}
        Here are some categories of details and describe aspects to focus on under the category, if the details extracted from the answer can not be classified into the following categories, you can think of your own: \texttt{\textbackslash n}
        1. People or Animals:  Describe their expressions, facial features, postures, clothing, age,  and quantity. Please include any notable actions or interactions they are involved in. \texttt{\textbackslash n}
        2. Plants: Describe the quantity, types, size, color, and any notable features such as flowers or fruits. \texttt{\textbackslash n}
        3. Food: Describe the quantity, types, colors, and presentation (e.g., plated, packaged). \texttt{\textbackslash n}
        4. Buildings:  Describe the quantity, types, architectural style, appearance, shapes, and any distinctive features (e.g., windows,  doors, decorations).
        5. Vehicles: Describe the types, appearance, quantity, color, and any notable features (e.g., brand, model, condition). \texttt{\textbackslash n}
        Please answer in this format: \texttt{\textbackslash n}
        Detail: [the detail what the model think itself should focus on in MODEL ANSWER.] \texttt{\textbackslash n}
        Category: [What category does this detail fall into. Can either pick from the above categories or thick by yourself.] \texttt{\textbackslash n}
        Relevant Describe Aspects: [The possible describe aspects when focusing on the detail.] \texttt{\textbackslash n}
        Here is the MODEL ANSWER: {model\_answer} \texttt{\textbackslash n}
        Output: \texttt{\textbackslash n} \\
        
        \midrule
        New Temporal Perspective Description & Please describe the video from a new temporal perspective. Focus on changes that occur before and after a specific camera transition or time point in the video.\texttt{\textbackslash n\textbackslash n}Here are the previously mentioned perspectives for reference:\texttt{\textbackslash n\{previous perspectives\}}\\
        \midrule
        New Spatial Perspective Description & Please describe the video from a new spatial perspective. Focus on different areas of the frame, such as the left side, right side, foreground, or background. \texttt{\textbackslash n\textbackslash n}Here are the previously mentioned perspectives for reference:\texttt{\textbackslash n\{previous perspectives\}}\\
        \midrule
        Background Description & Please provide a detailed description of the background in the video. Focus on the setting, environment, and any contextual elements that contribute to the overall scene. \texttt{\textbackslash n\textbackslash n}Here are the previously mentioned perspectives for reference:\texttt{\textbackslash n\{previous descriptions\}}\\
        \midrule
        Camera Move Description & Please describe the changes in camera shots and movements throughout the video. Focus on different types of shots, camera movements, angles, transitions, and any special effects used. \texttt{\textbackslash n\textbackslash n}Here are the previously mentioned perspectives for reference:\texttt{\textbackslash n\{previous descriptions\}}\\
        \bottomrule \\
    \end{tabularx}
    \caption{The action prompt used in ``expand'' phase of MCTS.}
    \label{tab: Action type and prompt}
\end{table*}

\subsection{MCTS Evaluation Phase Details}
\label{appendix: Evaluation Details}
In the evaluation phase of MCTS iteration, we need to calculate the MC value for each newly expanded node. The MC value of each node \(s\) depends on the key points contained within node \(s\). Therefore, we first need to extract the key points from node \( s \). We use GPT-4 to extract key points from the behavior results represented by the node, and the prompt used for extracting key points can be found in Figure \ref{fig: prompt-extract kp}. 

Next, we need to ask questions about the information of each extracted key point to facilitate subsequent verification work. As shown in Figure \ref{fig: kp_verify} in Section \ref{sec: evaluation}, each key point will generate n (\( n \geq 1 \)) yes/no questions. For each yes/no question, we use two powerful MLLM as verifiers to answer each question. A key point is only considered correct if all its questions pass verification in both verifiers. We use GPT-4 to generate questions, and the corresponding prompt can be found in Figure \ref{fig: prompt-generate questions}. During verification, both verifiers use the prompt displayed in Figure \ref{fig: prompt-answer verification questions}.

The evaluation phase is the most critical step in MCTS because the evaluation of each node influences subsequent backpropagation and the selection in the next MCTS interation. More specifically, the verification of each key point affects the MC value, which in turn impacts the evaluation results of each node. Below, we introduce other methods we have attempted to verify key points and the reasons why they were ultimately not adopted.

\begin{figure}
    \centering
    \includegraphics[width=\linewidth]{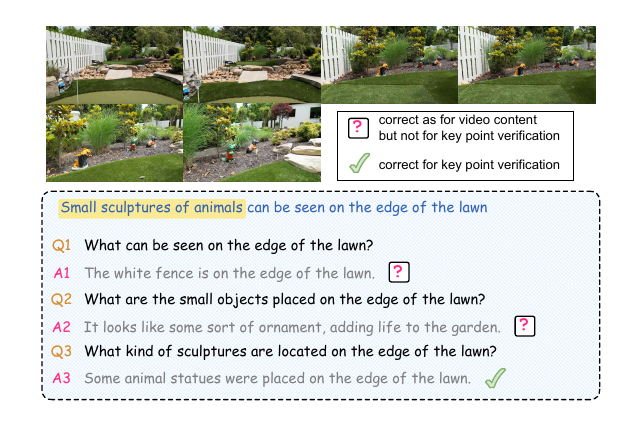}
    \caption{Illustration of ``Key-Info as Answer'' verification method.}
    \label{fig: kp_verify_other}
\end{figure}

\begin{enumerate}
    \item \textbf{Paraphrase + Multi-Verify}: We initially create three paraphrased versions of each key point extracted from the video and employed a Verifier to ascertain if each paraphrased point is indeed present in the video. The Verifier we used is Qwen2-VL-72B. However, during our experiments, it becomes evident that the Verifier has issues with misclassification. For instance, it mistakenly recognizes the number 5 on an athlete's jersey as 6 in all three paraphrased versions of the key point, resulting in incorrect information being included in the verified key points. Furthermore, we note that using paraphrasing as a means of multiple verification was not effective, since the Verifier often provide consistent assessments across the paraphrases of the same key point.
    \item \textbf{Key-Info as Answer}: In this approach, we first use GPT-4 to identify key information and then formulate three questions with the key information as the answer. We then use GPT-4o to answer these three questions, considering the key point correct if at least two out of the three answers matched the key information. However, most of the generated questions are open-ended, leading to non-unique answers for the same video. Figure \ref{fig: kp_verify_other} illustrates the drawbacks of this method. As shown in Figure 
    \ref{fig: kp_verify_other}, it can be found that the answers to the first two questions are consistent with the video content, but they cannot be used to judge correctness of the key point. Because the key point generated using MCTS are so detailed and subtle, the verifier cannot answer the information we want in such a open-ended question.
\end{enumerate}

Ultimately, our method improves upon Method 2 by modifying the questions to a yes/no format to ensure clear directionality. To enhance the accuracy of the verification, we select two powerful MLLMs (GPT-4o and Qwen2-VL-72B) based on the evaluation results from OpenCompass \footnote{\url{https://rank.opencompass.org.cn/home}. An open-source, efficient, and comprehensive large model evaluation system and open platform} by November.

\subsection{Post Processing Details}
\label{appendix: Post Processing Details}
During the evaluation phase of MCTS, we have already determined the correctness of the key points contained in each generated description. In the filtering stage, we need to perform two steps: 1. Filter out the correct key points in each node, 2. Use GPT-4 to filter out overly subjective and broad key points. These key points may be similar to the data distribution of the generator's output (Qwen2-VL-7B), leading to distorted evaluation results. Next, for the filtered key points, we remove duplicates, retaining a similarity of 0.9 to increase the diversity of key points by using ``all-MiniLM-L6-v2 ''~\footnote{\url{https://huggingface.co/sentence-transformers/all-MiniLM-L6-v2}}. To further address the issue of key point data distribution, we used GPT-4 to paraphrase all key points. However, this operation had almost no impact on the evaluation results, possibly because the data distribution correction was already performed when using GPT-4 to extract key points from the nodes. The prompt we used during filtering phase and paraphrasing phase can be found in Figure \ref{fig: prompt-filter kp} and Figure \ref{fig: prompt-paraphrase kp} respectively. As a benchmark, it is essential to ensure that the key points generated by AutoCaption are entirely reliable. Therefore, we let annotators check the key points generated under a similarity threshold of 0.9, retaining only those that accurately reflect the video content, with a human verification pass rate of 94.7\%.

\section{MCTS-VCB}
\subsection{MCTS-VCB Statistics}
\label{appendix: MCTS-VCB Statistics}
\begin{figure}[!ht]
    \centering
    \includegraphics[width=\linewidth]{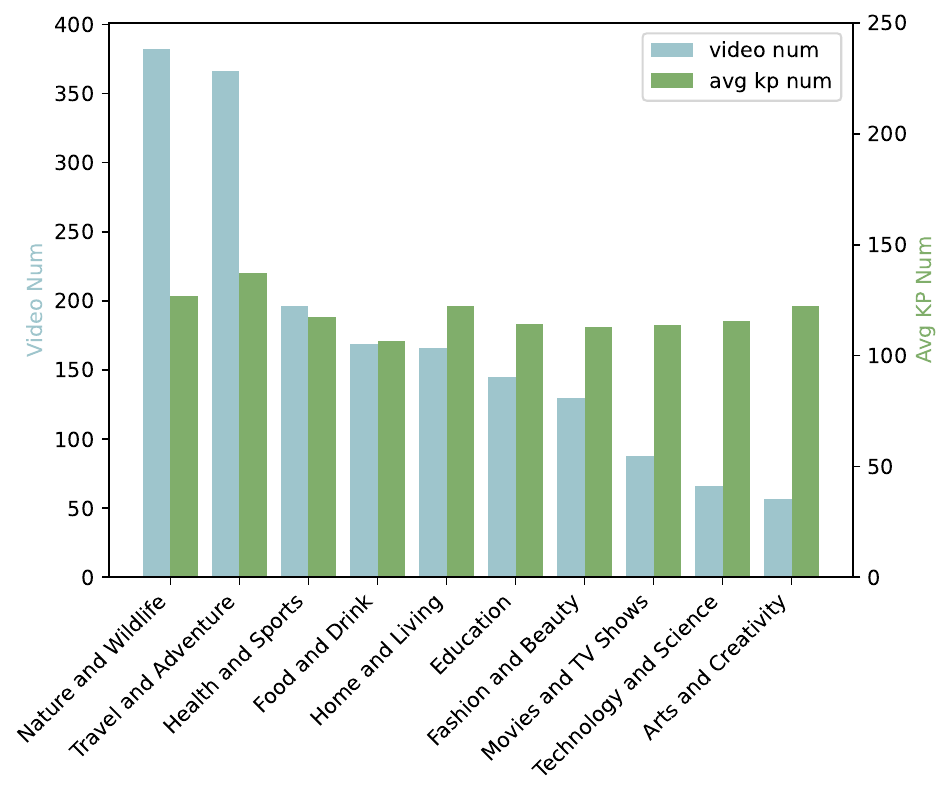}
    \caption{Video counts and average key point counts across different video categories.}
    \label{fig: video_category_bar}
\end{figure}

After the refined process of AutoCaption, we obtained an accurate and comprehensive key point data pool for each video, MCTS-VCB. MCTS-VCB comprises 1,765 videos, with an average length of 12.39 seconds per video and an average of 122.3 keypoints per video. The video categories are diverse, encompassing ten categories: 1) Movies and TV Shows, 2) Education, 3) Health and Sports, 4) Fashion and Beauty, 5) Home and Living, 6) Travel and Adventure, 7) Nature and Wildlife, 8) Technology and Science, 9) Arts and Creativity, and 10) Food and Drink. As shown in Figure \ref{fig: video_category_bar}, the \textit{Nature and Wildlife} category has the highest number of videos, reaching 382, while the Arts and Creativity category has the fewest, with 57 videos. Additionally, the average number of keypoints contained in videos of different categories does not vary significantly, which demonstrates the applicability of AutoCaption across various types of videos.

\begin{figure}[!ht]
    \centering
    \includegraphics[width=\linewidth]{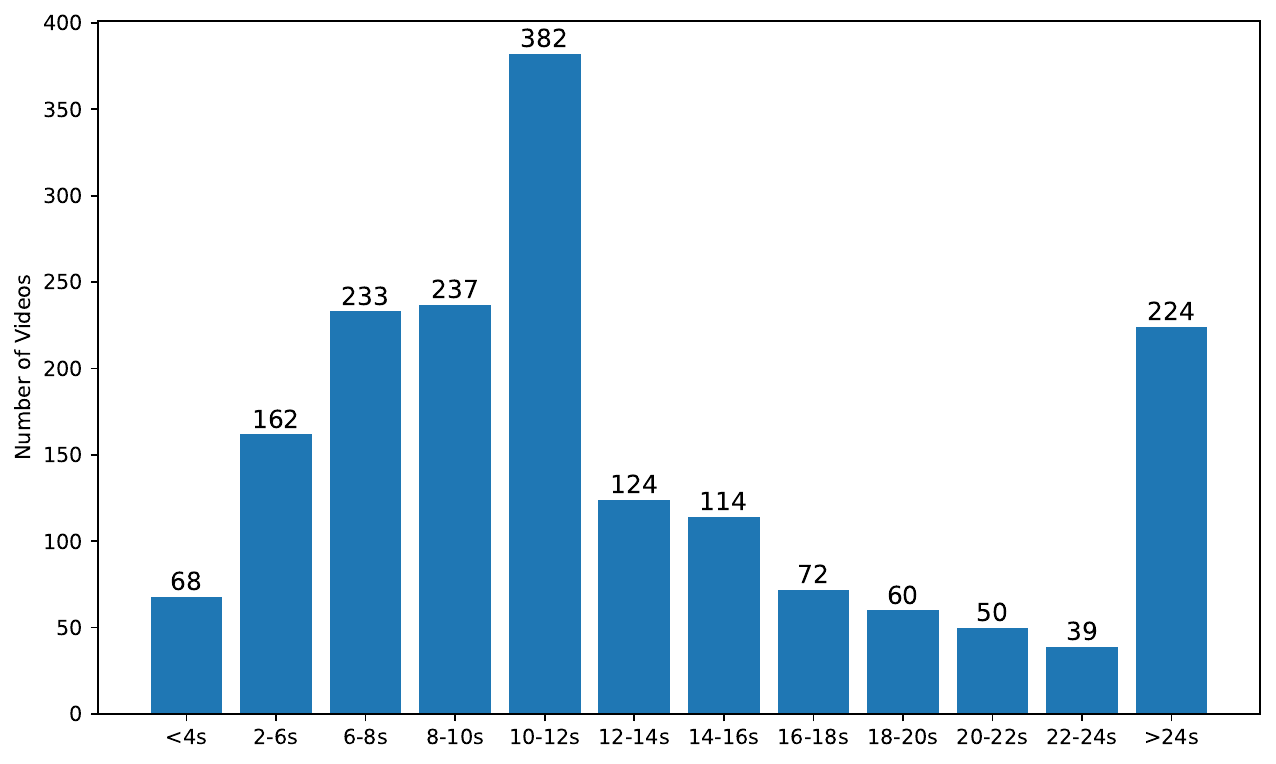}
    \caption{Video length distribution in seconds.}
    \label{fig:video_length_bar}
\end{figure}

According to Figure~\ref{fig:video_length_bar}, the highest number of videos falls within the 10–12s range, reaching 382. There is a reasonable distribution of videos across different duration ranges. With 224 videos exceeding 24s, this allows for a comprehensive evaluation of MLLMs' performance across various video lengths.

\begin{figure}[!ht]
    \centering
    \includegraphics[width=\linewidth]{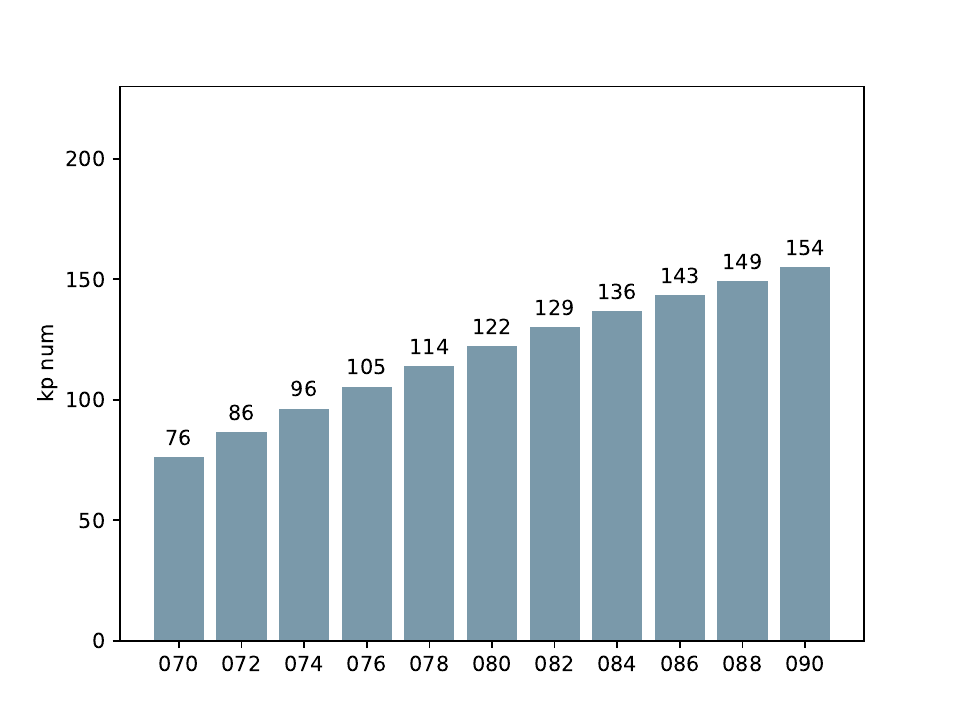}
    \caption{Video counts across different deduplicate similarity threshold.}
    \label{fig: kp_num_threshold}
\end{figure}

In AutoCaption, it is verified to obtain detailed, comprehensive, and nuanced key points of a video. However, when merging key points from different nodes \(s\) in search tree \(T\), semantically similar key points may exist, and the subtle differences contained within these semantically similar key points could be a weakness of MLLMs. Therefore, when forming MCTS-VCB, we also analyzed data with different similarity thresholds. We evaluated ten thresholds ranging from 0.7 to 0.9, and as shown in Figure \ref{fig: kp_num_threshold}, the average number of key points per video gradually increases with higher similarity thresholds. We also discuss the impact of different similarity thresholds on evaluation results in the experimental section \ref{section: Performance across different similarity threshold}.

\begin{figure*}[!ht]
    \centering
    \includegraphics[width=\linewidth]{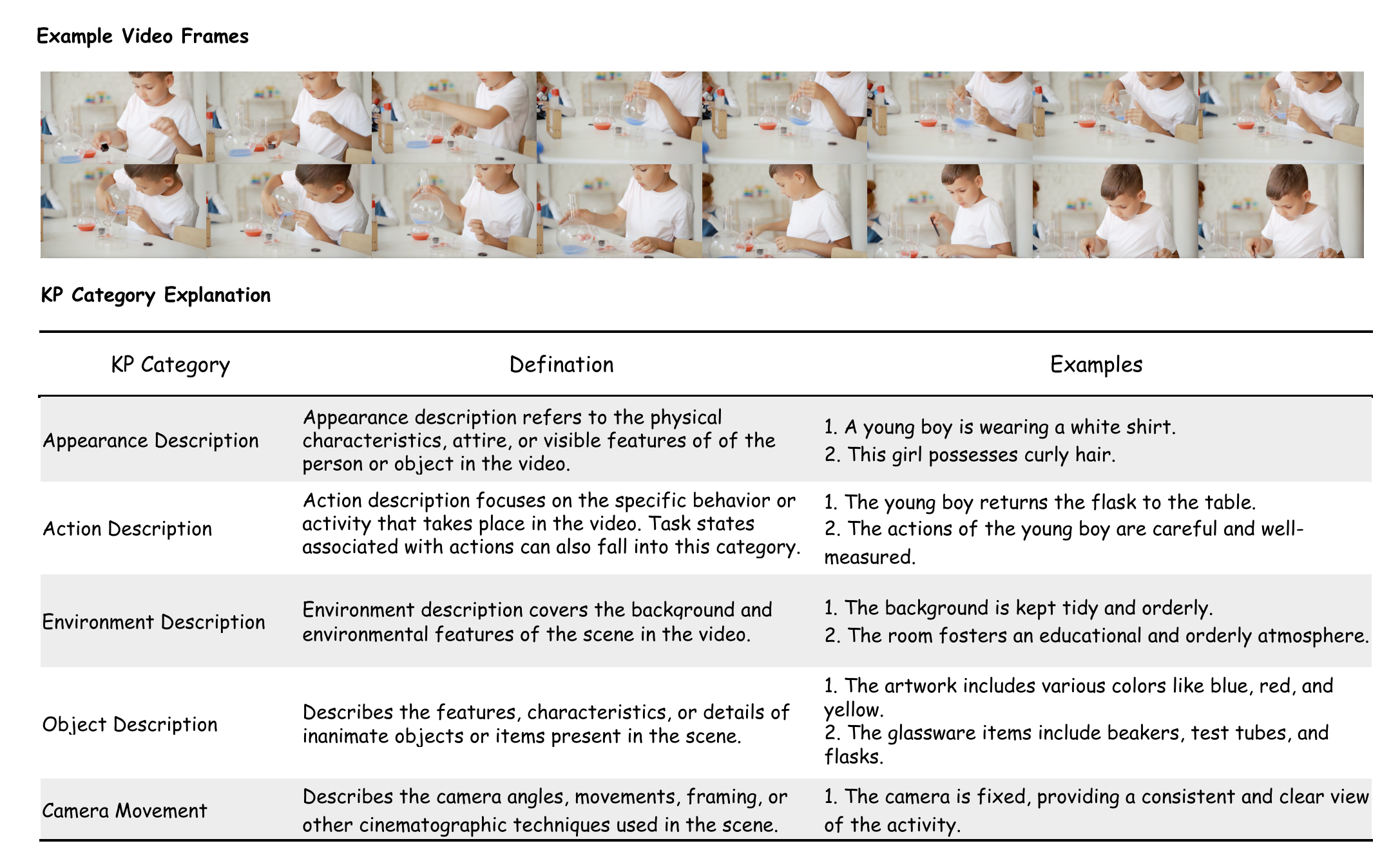}
    \caption{Illustration of key points categories.}
    \label{fig: MCTS-VCD-kp_category}
\end{figure*}

We categorized the key points into five types: 1) Appearance Description, 2) Action Description, 3) Environment Description, 5) Object Description, and 6) Camera Movement and Composition. The defination and examples of each category is described in Figure \ref{fig: MCTS-VCD-kp_category}.

\begin{figure}[!ht]
    \centering
    \includegraphics[width=\linewidth]{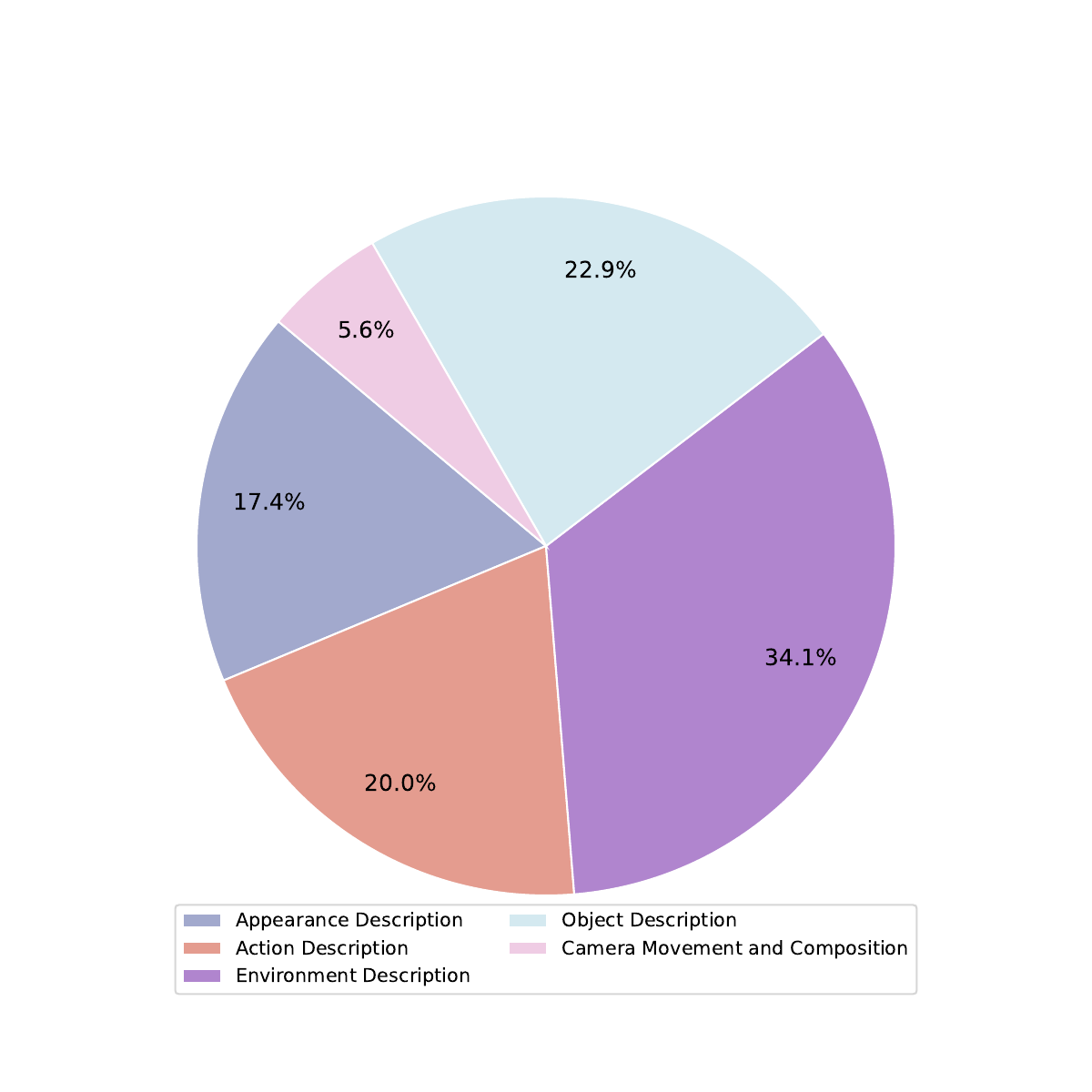}
    \caption{The distribution of key points of a video.}
    \label{fig: category_kp_count_per_video}
\end{figure}

Additionally, Figure \ref{fig: category_kp_count_per_video} shows the distribution of different keypoint categories in an average video. 
It can be observed that \textit{Environment Description} keypoint type is the most prevalent among the key points. However, there is currently no dataset that focuses on environment descriptions when evaluating video captioning tasks, which is an advantage of MCTS-VCB. The second most common keypoint type is \textit{Object Description}, indicating that AutoCaption effectively generates key points related to object details, surpassing other key point types such as \textit{Action Description} that are the focus of other works. 

\subsection{Evaluation Experiments Setting}
\label{appendix: experiment_setting}
We evaluate 21 open- and close-source MLLMs, with varying sizes. The model cards are shown in Table \ref{tab: model cards}. We used different prompts for different series of MLLMs, which can be categorized into the following three classes:
\begin{enumerate}
    \item For VILA series, we use the prompt displayed on GitHub \footnote{\url{https://github.com/NVLabs/VILA?tab=readme-ov-file\#video-captioning}}: ``Elaborate on the visual and narrative elements of the video in detail.\textquotedblright
    \item For Pllava series, we use the prompt shown on official script \footnote{\url{https://github.com/magic-research/PLLaVA/blob/main/tasks/eval/recaption/pllava_recaption.py\#L121C1-L129C6}}: ``You are to assist me in accomplishing a task about the input video. Reply to me with a precise yet detailed response. For how you would succeed in the recaptioning task, read the following Instructions section and Then, make your response with a elaborate paragraph.\texttt{\textbackslash n} \# Instructions\texttt{\textbackslash n} 1. Avoid providing over detailed information such as color, counts of any objects as you are terrible regarding observing these details\texttt{\textbackslash n} 2. Instead, you should carefully go over the provided video and reason about key information about the overall video\texttt{\textbackslash n} 3. If you are not sure about something, do not include it in you response.\texttt{\textbackslash n} \# Task\texttt{\textbackslash n} Describe the background, characters and the actions in the provided video.\texttt{\textbackslash n}\textquotedblright
    \item For all other MLLMs, we use ``Please describe the video in detail.\textquotedblright
\end{enumerate}

\subsection{Supplementary Experiment Results}
\label{appendix: Supplementary Experiment Results}
\begin{figure*}[!ht]
    \centering
    \includegraphics[width=\linewidth]{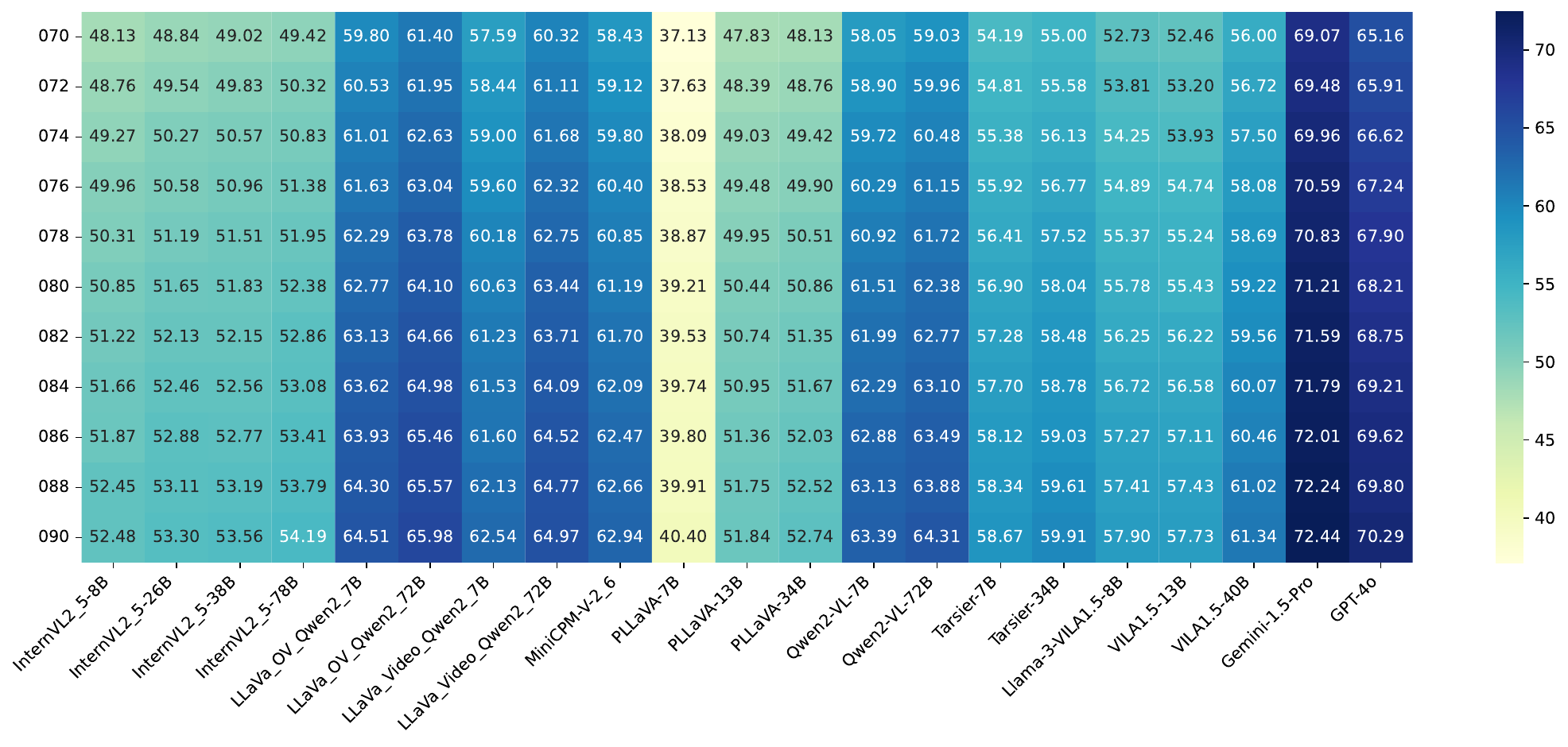}
    \caption{F1 score across different similarity threshold.}
    \label{fig: f1_score_across_threshold_heatmap}
\end{figure*}

As shown in Figure \ref{fig: f1_score_across_threshold_heatmap}, it is interesting to note that as the similarity threshold increases, the evaluation scores of the models also increase. This phenomenon is reasonable because the final key points, after deduplication using higher similarity thresholds, contain more similar key points. These key points are more easily generated by the generator during MCTS generation. Upon observation, we found that these key points and the similar parts of the descriptions for the same video by different models are generally consistent. This consistency leads to better performance in the evaluation sets filtered by higher similarity thresholds. To ensure that MCTS-VCB includes key points with a certain degree of semantic similarity (since semantic similarity does not imply complete consistency in detailed information), all experimental results in this paper are based on a similarity threshold of 0.8.

\section{Training Data Curation Detail \& Example}
\label{appendix: Training Data Curation Detail  Example}
The construction of the training data followed AutoCaption but omitted the human check in data post-processing and the ``Overall Description'' action is also executed by the generator. First, key points were obtained from a video within AutoCaption. Then, we used the prompt shown in Figure \ref{fig: prompt-gpt4o_generate_thought_process} to guide GPT-4 in merging the key points into a thought process according to the generated order. As illustrated in Figure \ref{fig: prompt-gpt4o_generate_caption_based_on_thougt_process}, we placed the thought process within \texttt{ <thought>...</thought>} to guide GPT-4o in generating detail-focused captions based on the video and the thought process. The generated captions will be used as fine-tuning data.
\section{Complete Experimental Results}
\label{appendix: Complete Experimental Results}
The specific result of the fine-tuned MLLM's performance on different dimensions of MCTS-VCB and DREAM-1K are displayed in Table \ref{tab: full_res_performance_improvement_after_sft}.

The complete result of precision, recall and F1 score of MLLMs' performance across different video categories is displayed in Table \ref{tab: all_result_video_category}.

The complete result of precision, recall and F1 score of MLLMs' performance across different similarity thresholds used in merging key points from all nodes \(s\) of one search tree \(T\) is displayed in Table \ref{tab: all_result_threshold}.

\begin{table*}[!t]
    \centering
    \scriptsize
    \begin{tabularx}{0.8\linewidth}{ 
			>{\raggedright\arraybackslash\hsize=1\hsize\linewidth=\hsize}X
			>{\centering\arraybackslash\hsize=1\hsize\linewidth=\hsize}X
			>{\centering\arraybackslash\hsize=1.2\hsize\linewidth=\hsize}X
			>{\centering\arraybackslash\hsize=0.8\hsize\linewidth=\hsize}X
		}
        \toprule
        {Model Name} & Language Model & Vision Model & \#Frames \\
        \midrule
\rowcolor{tabelcellgrey} \multicolumn{4}{l}{\textbf{\textit{Open-Source MLLMs}}} \\
InternVL2\_5-8B &   Internlm2\_5-7b-chat &  InternViT-300M-448px-V2\_5 & 16  \\
InternVL2\_5-26B &   Internlm2\_5-20b-chat &  InternViT-6B-448px-V2\_5 & 16  \\
InternVL2\_5-38B &   Qwen2.5-32B &  InternViT-6B-448px-V2\_5 & 16  \\
InternVL2\_5-78B &  Qwen2.5-72B &  InternViT-6B-448px-V2\_5 & 16  \\
LLaVa\_OV\_Qwen2\_7B &   Qwen2-7B &  SigLIP-SO400M & 16  \\
LLaVa\_OV\_Qwen2\_72B &   Qwen2-72B &  SigLIP-SO400M & 16  \\
LLaVa\_Video\_Qwen2\_7B &  Qwen2-7B &  SigLIP-SO400M & 16  \\
LLaVa\_Video\_Qwen2\_72B & Qwen2-72B &  SigLIP-SO400M & 16  \\
MiniCPM-V-2\_6 &  Qwen2-7B &  SigLIP-SO400M & 16  \\
PLLaVA-7B &  Vicuna-7B-v1.5 &  CLIP ViT & 16  \\
PLLaVA-13B &  Vicuna-13B-v1.5 &  CLIP ViT & 16  \\
PLLaVA-34B & Nous-Hermes-2-Yi-34B &  CLIP ViT & 16  \\
Qwen2-VL-7B-Instruct &  Qwen2-7B & ViT-675M & 16  \\
Qwen2-VL-72B-Instruct & Qwen2-72B & ViT-675M & 16  \\
Tarsier-7B &  Vicuna-7B-v1.5 & CLIP ViT & 16  \\
Tarsier-34B & Nous-Hermes-2-Yi-34B & CLIP ViT & 16  \\
Llama-3-VILA1.5-8B &  Llama3-8B & SigLIP-SO400M & 16  \\
VILA1.5-13B & Vicuna-7B-v1.5 & SigLIP-SO400M & 16  \\
VILA1.5-40B & vicuna1.5 & SigLIP-SO400M & 15  \\
\rowcolor{tabelcellgrey} \multicolumn{4}{l}{\textbf{\textit{Closed-Source MLLMs}}} \\
Gemini-1.5-Pro&  / & / & 16  \\
GPT-4o &  / & /& 16  \\
        \bottomrule
    \end{tabularx}
    \caption{The model cards of 21 evaluated MLLMs, including both open- and close-source ones. For most MLLMs, we sample 16 frames uniformly to represent videos; however, for VILA1.5-40B, due to context length limits, we sample 15 frames uniformly as input.}
    \label{tab: model cards}
\end{table*}

\begin{figure*}[!h]
\centering
\begin{tcolorbox}[colback=gray!10!white, colframe=black]
 Please evaluate the given video according to the given video content and corresponding caption to determine if it meets the following criteria: \\

1. The video must have a clear subject (e.g., person, animal, object, etc.). \\
2. The video should not contain a lot of special effects. \\
3. The video cannot contain long still clips. \\
4. The video must not be confused, unclear meaning. \\ \\

If the video meets all the criteria, output [yes]; if it fails to meet any of the criteria, output [no]. The input format is:
``Judgment: [yes/no] Reason: [Brief explanation]'' \\

Please provide your judgment and briefly explain the reason.
\end{tcolorbox}
\caption{The prompt used for GPT-4o to filter video clips.}
\label{fig: prompt-filter video}
\end{figure*}

\begin{figure*}[!t]
\centering
\begin{tcolorbox}[colback=gray!10!white, colframe=black]
    Please break down the following video caption into individual key points. A sentence in the video caption should be split to individual key points if it cantains much information. \\

    Requirement: \\
    1. Make sure to substitute pronouns in split individual sentences by the nouns they refer to. \\
    2. If a individual sentence contains uncertain expressions, these expressions should not be included as key points. \\

    For clarity, consider these examples: \\

    \#\# Example 1 \\
    \#\#\# Video Description:  \\
    \texttt{The video showcases ...} \\
    \#\#\# Result: \\
    > \texttt{\{Atomic key point\}} \\
    > \texttt{\{Atomic key point\}} \\
    > \texttt{...} \\

    \textcolor{myblue}{\{two more examples\}} \\

    With these examples in mind, please help me break down the following video caption into individual key points.
    \\

    \#\#\# Video Description:  \\
    {description} \\
    \#\#\# Result: 
\end{tcolorbox}
\caption{The prompt used for Qwen2-VL-72B to extract key points}
\label{fig: prompt-extract kp}
\end{figure*}

\begin{figure*}[!t]
\centering
\begin{tcolorbox}[colback=gray!10!white, colframe=black]
Please objectively classify the relationship between each video caption breakdown and provided human-generated key points.\\
Analyze each breakdown point individually to determine its relationship with the human-generated key points.\\

For each breakdown point, classify the relationship into one of the following categories:\\
1. ``entailment '' means that the breakdown point is accurately reflected within one or more of the human-generated key points.\\
2. ``contradiction '' means that breakdown point some detail in the breakdown point contradicts with the infomation mentioned human-generated key points.\\
3. ``neutral '' means that the relationship is neither ``entailment '' nor ``contradiction \textquotedblright.\\

For each breakdown point, provide a brief analysis explaining the reasoning behind your judgment.\\

Please present the result in a JSON dict format: \texttt{\{``
 breakdown\_point\_1'': \{`` judgement'': judgement\_1, `` analysis'': analysis\_1\}, ...,  ``
 breakdown\_point\_n'': \{`` judgement'': judgement\_n, `` analysis'': analysis\_n\}\}}.\\

For clarity, consider these examples:\\

\textcolor{myblue}{\{two examples\}} \\

With these examples in mind, please help me evaluate whether each breakdown point is accurately reflected in the provided human-generated key points.\\

\#\#\# Human-Generated Video Key Points:\\
{key\_point}\\

\#\#\# LMM Caption Breakdown Point to Evaluate:\\
{caption}\\

\#\#\# Result:
\end{tcolorbox}
\caption{The prompt used for Qwen2.5-72B-Instruct to determine the relationship between \(KP_{model}\) and \(KP_{ref}.\) When to determine the relationship between \(KP_{ref}\) and \(D_{mdoel}.\), we provide \(D_{mdoel}.\) instead of ``LMM Caption Breakdown Point to Evaluate '' in this prompt.}
\label{fig: prompt-determine_entail_contra_netural}
\end{figure*}

\begin{figure*}[!t]
\centering
\begin{tcolorbox}[colback=gray!10!white, colframe=black]
You are an experienced linguist, please help me rewrite the \texttt{{sentence num}} sentences, requiring the semantics to remain unchanged, and not adding or removing any details. \\
All rewrited sentences should be put in a list. Please only output the list! \\

\#\# Sentences:  \\
\texttt{[``1. sentence1'',``2. sentence2'',``3. sentence3'']} \\
\#\# Output: \\
\texttt{[``1. rewrited sentence1'',``2. rewrited sentence2'',``3. rewrited sentence3'']} \\

\#\# Sentences: \\
\texttt{sentences} \\
\#\# Output:

\end{tcolorbox}
\caption{The prompt used for GPT-4 to paraphase key points.}
\label{fig: prompt-paraphrase kp}
\end{figure*}

\begin{figure*}[!t]
\centering
\begin{tcolorbox}[colback=gray!10!white, colframe=black]
    You are given a key point extracted from a video description. \\
    Your task is to identify the key information within the given key point and construct yes/no questions according to the identified key information. \\
    If the key point has more than one key information, you need to generate more than one question.  \\
    
    For clarity, consider these examples: \\
    \#\# Example 1 \\
    \#\#\# Key Point  \\
    \texttt{key point 1} \\
    \#\#\# Output \\
    > \texttt{question 1} \\
    > \texttt{...} \\

    \textcolor{myblue}{\{two more examples\}} \\

    With these examples in mind, please help me break down the following video caption into individual key points.
    \\

    \#\#\# Key Point  \\
    {key point} \\
    \#\#\# Output 
\end{tcolorbox}
\caption{The prompt used for Qwen2-VL-72B to generate verification questions of a key point.}
\label{fig: prompt-generate questions}
\end{figure*}

\begin{figure*}[!t]
\centering
\begin{tcolorbox}[colback=gray!10!white, colframe=black]
    You are provided with several frames extracted from a video and \texttt{ques\_num} yes/no questions related to the content of the video. \\
    Your task is to analysis the frames carefully and provide a clear and concise judgement (yes/no) to each question based on the video frames. \\
    For, each question, please provide a single sentence answer (format like ``[Judgement], [Reason]'') to the question based on the video frames' content. \\
    All answers should be put in a list.  \\
    Please only output the list! \\
    
    \#\# Example 1 \\
    \#\#\# Questions  \\
    \texttt{\{1. question 1\}} \\
    \texttt{\{2. question 2\}} \\
    \#\#\# Output \\
    \texttt{[``1. [Judgement 1], [Reason 1]'', ``2. [Judgement 2], [Reason 2]'']} \\

    \textcolor{myblue}{\{one more example\}} \\

    \#\#\# Questions  \\
    \texttt{\{question list\}} \\
    \#\#\# Output 
\end{tcolorbox}
\caption{The prompt used for both GPT-4o and Qwen2-VL-72B to answer verification questions of a key point.}
\label{fig: prompt-answer verification questions}
\end{figure*}

\begin{algorithm*}
\caption{AutoCaption Workflow} 
\label{alg:autocaption} 
\begin{algorithmic}[1]
\State \textbf{Input:} 64 video frames uniformly sampled from video $v$.
\State \textbf{Output:} Final video key points

\State Initialize search tree $T$ with root node containing video $v$
\For{iteration $i = 1$ to 25}
    \State \textbf{Selection:} Select node $s_i$ with highest value using PUCT
    \State \textbf{Expansion:} Expand node $s_i$ with possible actions
    \State \textbf{Evaluation:} Calculate state value $Q(s, a)$ for new nodes
    \State \textbf{Backpropagation:} Update $N(s)$ and $Q(s, a)$ for nodes in $t(s_i)$
\EndFor

\State \textbf{Data Post-processing:}
\State \quad \textbf{Filter:} Remove irrelevant descriptions.
\State \quad \textbf{Deduplicate:} Eliminate duplicate descriptions.
\State \quad \textbf{Paraphrase:} Rephrase descriptions to alleviate data distribution of the generator.
\State \quad \textbf{Human Check:} Double check final key points.

\State \textbf{Return} Final key points of the video
\end{algorithmic}
\end{algorithm*}

\begin{figure*}[!t]
\centering
\begin{tcolorbox}[colback=gray!10!white, colframe=black]
You are given a list of key points describing a video. \\
Your task is to filter out key points that are too subjective, too minor, too general, or express speculation. \\
Additionally, any key points related to the historical background or cultural context of the video should also be filtered out, as they are not directly relevant to the video content. \\

For each key point, output the judgement ``yes'' if it should be kept, or ``no'' if it should be filtered out. Also output a reason for your judgement. \\

Criteria for filtering:\\
1. Subjective: Key points that rely heavily on personal feelings or interpretations.\\
2. Minor: Key points that are overly detailed and do not add significant value to the overall description.\\
3. General: Key points that are too broad and do not provide specific information about the video.\\
4. Speculative: Key points that express guesses or assumptions rather than concrete information.\\
5. Historical/Cultural: Key points that relate to the historical background or cultural context of the video.\\

For clarity, consider these examples: \\
    
    \#\# Example 1 \\
    \#\#\# Key Points  \\
    \texttt{\{1. These people could be tourists or travelers visiting the site.\}} \\
    \texttt{\{...\}} \\
    \#\#\# Output \\
    \texttt{1. ``[No] Speculative.\textquotedblright} \\
    \texttt{...} \\

    \textcolor{myblue}{\{two more examples\}} \\

    With these examples in mind, please help me filter out key points that are too subjective, too minor, too general, or express speculation.

    \#\#\# Key Points  \\
    \texttt{\{kep points\}} \\
    \#\#\# Output 
\end{tcolorbox}
\caption{The prompt used for GPT-4 to filter too subjective and broad key points.}
\label{fig: prompt-filter kp}
\end{figure*}

\begin{figure*}[!t]
\centering

\begin{tcolorbox}[colback=gray!10!white, colframe=black]
You are tasked with generating a structured internal thought process that explains how you analyze and describe a video. The goal is to demonstrate a step-by-step reasoning process, gradually refining the observations based on provided key points.\\

Formatting Requirements:\\
1. Your thought process should be enclosed within \texttt{<thought> ... </thought>} tags.\\
2. Your thought should contain the provided ``Overall Description'' and all (Observation,Key Point) pairs.\\
3. The reasoning should be sequential and structured, mimicking a natural process of observation and refinement.\\

\#\# Example Input and Output: \\
\#\#\# Input: \\
Overall Description:  
\texttt{\{overall\_description\}}\\

Observation: Vehicles  \\
Key Point: There is no existence of any vehicles in the video.\\

Observation: Buildings  \\
Key Point: None  \\

Observation: Snow-covered mountains and a sea of clouds  \\
Key Points:  \\
1. The mountain peaks stand tall against a backdrop of a sea of clouds.\\
2. The mountains appear in various shapes and sizes.\\
3. The mountains create a dynamic and picturesque landscape. \\
...\\

\texttt{\{more Observation-Key Point pairs\}}\\

\#\#\# Expected Output:\\
<thought>\texttt{\{thought process\}}</thought>\\

Remember the above example and requirements. Now, please help me integrate the following 'Overall Description' and multiple 'Observation - Key Point' pairs into a complete internal thought process.\\

\#\#\# Input:\\
Overall Description:  \\
\textcolor{myblue}{\{overall\_description\}}\\

\textcolor{myblue}{\{observation\_kp\_pairs\}}\\

\#\#\# Expected Output:
\end{tcolorbox}
\caption{The prompt used for GPT-4 to generate thought process according to video key points.}
\label{fig: prompt-gpt4o_generate_thought_process}
\end{figure*}

\begin{figure*}[!t]
\centering

\begin{tcolorbox}[colback=gray!10!white, colframe=black]
According to the reasoning thought, describe the video in detail. \\
\texttt{<thought>}
As I begin to analyze the video, I first take in the overall description to set the context. The scene is festive and joyful, featuring a young child sitting on Santa's lap in front of a Christmas tree. Santa is reading from an old-fashioned typewriter, and behind them, elves are busy preparing presents. This setting captures the essence of the holiday spirit, emphasizing the joy of giving and receiving gifts. \\

Moving on to specific observations, I notice the child's interaction with the typewriter. This typewriter is not just any typewriter; it's vintage, with a mechanical design that includes visible gears and levers. The body of the typewriter shows signs of wear and age, suggesting it has been cherished over many years. The child, pressing the keys and focused on their task, adds a personal touch to the scene. The surrounding environment complements this interaction, with a table filled with wrapped gifts and papers, and decorative elements in the background enhancing the festive atmosphere.\\

Next, I expand my focus to include other key elements in the scene. Central to the image is Santa Claus, seated at a table adorned with neatly wrapped presents. Santa, in his traditional red and white costume with a lush white beard, creates a classic holiday image. Beside him, the child in a bright yellow outfit adds a pop of color. The elves, dressed in green uniforms with red hats, are a testament to the busy and cheerful environment. The table not only holds the typewriter but also various other items that contribute to the festive setting.\\

The background elements also play a significant role in enriching the scene. A beautifully decorated Christmas tree stands tall, its branches laden with twinkling lights and adorned with ornaments, creating a magical backdrop. Additionally, a large window to the side of the tree allows natural light to flood the room, enhancing the warmth and inviting nature of the setting.\\

By examining these details, from the child's interaction with the typewriter to the broader festive environment, I gain a comprehensive understanding of the video. Each element, whether it's the vintage typewriter, Santa's traditional attire, or the elves' busy preparations, contributes to the overall holiday spirit captured in this scene. The video beautifully portrays the magic of the holiday season through these meticulously arranged and detailed festive elements.\\
\texttt{</thought>}
\end{tcolorbox}
\caption{An example of prompting GPT-4o to generate more detailed caption based on thought process.}
\label{fig: prompt-gpt4o_generate_caption_based_on_thougt_process}
\end{figure*}

\begin{table*}[!t]
    \centering
    \scriptsize
    \begin{tabularx}{\linewidth}{ 
        >{\raggedright\arraybackslash\hsize=1\hsize\linewidth=\hsize}X
        >{\centering\arraybackslash\hsize=1\hsize\linewidth=\hsize}X
        >{\centering\arraybackslash\hsize=1\hsize\linewidth=\hsize}X
        >{\centering\arraybackslash\hsize=1\hsize\linewidth=\hsize}X
        >{\centering\arraybackslash\hsize=1\hsize\linewidth=\hsize}X
        >{\centering\arraybackslash\hsize=1\hsize\linewidth=\hsize}X
    }
        \toprule
        MCTS-VCB & \makecell{Appearance \\ Description} & \makecell{Action \\ Description} & \makecell{Environment \\ Description} & \makecell{Object \\ Description} & \makecell{Camera Movement \\ and Composition} \\
        \midrule
        Vanilla &  65.0/42.0/51.0 &  76.0/48.7/59.4 &  72.1/30.3/42.7 &  54.7/31.7/40.1 &  69.6/40.1/50.8  \\
        Fine-tuned & 74.2/48.6/58.7 &  78.6/59.9/68.0 &  81.0/62.5/70.5 &  74.1/43.9/55.1 &  72.0/32.3/44.6  \\
        \midrule
        DREAM-1K & Movie Animation & Movie Live Action & Shorts & Stock & Youtube \\
        \midrule
        Vanilla &  28.1/16.1/20.5 & 30.1/25.0/27.3 & 39.5/24.2/30.0 & 37.8/31.6/34.4 & 36.4/26.7/30.8 \\
        Fine-tuned &  31.1/25.2/27.8 & 33.9/31.6/32.7 & 40.7/28.1/33.2 & 38.6/37.7/38.2 & 39.5/31.1/34.8 \\
        \bottomrule
    \end{tabularx}
    \caption{The performance improvement across different dimensions of MCTS-VCB and DREAM-1K. Each cell contains ``\textbf{Precision} / \textbf{Recall} / \textbf{F1 Score}\textquotedblright.}
    \label{tab: full_res_performance_improvement_after_sft}
\end{table*}

\begin{table}
\footnotesize
\rotatebox{90}{\begin{minipage}[b][\textwidth][c]{\textheight}
\centering
    \begin{tabularx}{\textheight}{ 
			>{\raggedright\arraybackslash\hsize=2\hsize\linewidth=\hsize}X
			>{\centering\arraybackslash\hsize=0.9\hsize\linewidth=\hsize}X
			>{\centering\arraybackslash\hsize=0.9\hsize\linewidth=\hsize}X
			>{\centering\arraybackslash\hsize=0.9\hsize\linewidth=\hsize}X
			>{\centering\arraybackslash\hsize=0.9\hsize\linewidth=\hsize}X
			>{\centering\arraybackslash\hsize=0.9\hsize\linewidth=\hsize}X
			>{\centering\arraybackslash\hsize=0.9\hsize\linewidth=\hsize}X
			>{\centering\arraybackslash\hsize=0.9\hsize\linewidth=\hsize}X
			>{\centering\arraybackslash\hsize=0.9\hsize\linewidth=\hsize}X
			>{\centering\arraybackslash\hsize=0.9\hsize\linewidth=\hsize}X
			>{\centering\arraybackslash\hsize=0.9\hsize\linewidth=\hsize}X
		}
        \toprule
        {Model Name} & Movies and TV Shows & Education & Health and Sports & Fashion and Beauty & Home and Living & Travel and Adventure & Nature and Wildlife & Technology and Science & Arts and Creativity & Food and Drink \\
        \midrule
\rowcolor{tabelcellgrey} \multicolumn{11}{l}{\textbf{\textit{Open-Source MLLMs}}} \\
InternVL2\_5-8B & 69.8/39.7/50.6 &  71.0/40.9/51.9 &  66.9/42.3/51.8 &  68.6/39.7/50.3 &  69.7/39.1/50.1 &  68.7/40.5/50.9 &  71.5/47.2/56.9 &  69.8/42.7/53.0 &  73.2/44.8/55.6 &  67.7/40.9/51.0     \\
InternVL2\_5-26B & 71.3/41.4/52.4 &  72.1/42.2/53.3 &  67.7/43.1/52.7 &  69.1/40.5/51.0 &  69.2/38.8/49.8 &  69.8/41.5/52.1 &  71.5/47.9/57.4 &  67.5/44.1/53.4 &  72.7/43.9/54.8 &  70.0/42.3/52.7     \\
InternVL2\_5-38B & 70.5/40.1/51.2 &  72.1/42.8/53.7 &  69.5/43.7/53.6 &  70.5/40.1/51.1 &  70.8/41.0/51.9 &  70.5/41.1/51.9 &  71.7/47.1/56.9 &  68.8/44.7/54.2 &  69.5/44.6/54.3 &  68.9/43.7/53.5    \\
InternVL2\_5-78B & 72.7/42.1/53.3 &  72.3/41.8/53.0 &  69.9/44.7/54.6 &  71.0/41.3/52.2 &  69.6/40.3/51.1 &  70.5/42.5/53.0 &  70.9/48.1/57.3 &  68.6/44.3/53.9 &  73.0/46.6/56.8 &  69.2/43.2/53.2     \\
LLaVa\_OV\_Qwen2\_7B & 77.3/52.8/62.7 &  77.7/55.1/64.5 &  78.0/55.0/64.5 &  77.9/52.4/62.6 &  77.2/52.5/62.5 &  80.0/52.2/63.2 &  83.2/56.4/67.2 &  79.7/53.6/64.1 &  82.6/57.0/67.5 &  78.4/55.2/64.8     \\
LLaVa\_OV\_Qwen2\_72B & 78.8/51.7/62.4 &  77.9/56.4/65.4 &  78.4/57.2/66.1 &  79.5/52.0/62.9 &  79.1/53.9/64.1 &  82.1/52.5/64.0 &  85.1/58.1/69.1 &  84.7/53.9/65.9 &  82.3/58.4/68.3 &  80.2/57.2/66.8     \\
LLaVa\_Video\_Qwen2\_7B & 78.4/47.2/59.0 &  78.9/49.7/61.0 &  78.7/50.6/61.6 &  78.7/47.2/59.0 &  78.1/48.7/60.0 &  79.9/50.3/61.8 &  81.7/55.6/66.2 &  79.6/50.7/62.0 &  82.0/56.3/66.7 &  79.1/52.4/63.0    \\
LLaVa\_Video\_Qwen2\_72B & 80.6/52.1/63.3 &  80.3/53.9/64.5 &  81.4/55.3/65.9 &  79.9/51.7/62.8 &  79.4/52.5/63.2 &  83.1/51.6/63.7 &  85.0/56.8/68.1 &  82.4/55.0/66.0 &  80.9/57.0/66.8 &  81.2/55.3/65.8     \\
MiniCPM-V-2\_6 & 74.9/50.8/60.5 &  76.4/49.4/60.0 &  78.1/53.0/63.2 &  76.0/49.6/60.1 &  75.6/49.2/59.6 &  79.3/52.3/63.0 &  79.2/56.7/66.1 &  78.0/51.2/61.8 &  80.3/56.2/66.1 &  75.9/51.0/61.0     \\
PLLaVA-7B & 58.4/33.6/42.7 &  50.8/26.8/35.1 &  59.7/33.6/43.0 &  57.6/33.7/42.5 &  53.2/28.4/37.0 &  58.7/32.9/42.1 &  54.5/36.2/43.5 &  43.9/22.8/30.0 &  49.7/28.3/36.1 &  50.3/26.9/35.1    \\
PLLaVA-13B & 73.9/39.5/51.5 &  71.5/38.9/50.4 &  70.1/39.6/50.6 &  70.8/38.5/49.9 &  71.7/38.1/49.8 &  73.0/39.3/51.1 &  75.1/45.9/56.9 &  73.3/39.8/51.6 &  75.7/44.4/56.0 &  68.3/37.2/48.2      \\
PLLaVA-34B & 72.0/39.7/51.2 &  75.3/38.5/51.0 &  74.9/41.0/53.0 &  71.5/38.7/50.3 &  74.7/38.3/50.7 &  74.5/38.6/50.8 &  78.0/45.4/57.4 &  76.2/41.2/53.5 &  78.3/44.4/56.7 &  71.3/37.8/49.4     \\
Qwen2-VL-7B & 76.0/49.8/60.1 &  75.6/52.5/62.0 &  76.2/54.0/63.2 &  75.3/52.0/61.5 &  76.8/51.3/61.5 &  78.7/51.8/62.5 &  81.1/56.9/66.9 &  75.2/54.0/62.8 &  79.4/56.7/66.2 &  75.8/54.9/63.7    \\
Qwen2-VL-72B & 75.9/51.7/61.5 &  76.2/54.4/63.5 &  73.8/55.6/63.5 &  74.1/52.4/61.4 &  76.5/52.9/62.5 &  80.0/53.0/63.7 &  81.4/58.3/67.9 &  75.7/56.0/64.3 &  78.9/55.4/65.1 &  74.7/55.2/63.5     \\
Tarsier-7b & 72.4/45.0/55.5 &  73.7/46.9/57.3 &  74.7/47.2/57.9 &  73.3/44.8/55.6 &  73.2/44.7/55.5 &  80.8/46.4/58.9 &  81.3/52.6/63.9 &  75.6/47.7/58.5 &  76.6/48.9/59.7 &  70.8/44.1/54.4   \\
Tarsier-34b & 74.1/47.6/58.0 &  75.7/45.6/56.9 &  78.7/47.7/59.4 &  76.2/44.7/56.3 &  77.9/47.0/58.6 &  84.0/46.8/60.1 &  83.2/52.6/64.4 &  77.8/46.8/58.4 &  82.7/50.0/62.3 &  73.5/45.7/56.4    \\
Llama-3-VILA1.5-8b & 75.2/42.9/54.7 &  73.0/44.3/55.2 &  73.5/47.4/57.7 &  74.7/42.7/54.3 &  73.2/43.3/54.4 &  76.4/45.0/56.7 &  77.2/51.7/61.9 &  76.8/46.4/57.9 &  81.2/50.3/62.1 &  75.5/47.5/58.3    \\
VILA1.5-13b & 69.5/43.6/53.6 &  70.7/45.5/55.3 &  72.1/47.8/57.5 &  72.3/44.5/55.1 &  72.5/43.7/54.5 &  75.8/45.0/56.4 &  76.6/51.9/61.9 &  72.4/45.3/55.8 &  75.9/50.4/60.6 &  72.9/45.6/56.1    \\
VILA1.5-40b & 78.1/48.9/60.1 &  76.6/49.1/59.9 &  77.6/52.5/62.6 &  76.5/48.5/59.4 &  75.5/47.6/58.4 &  80.3/47.7/59.9 &  79.0/53.7/63.9 &  75.2/48.2/58.8 &  84.9/53.2/65.4 &  77.3/50.6/61.2    \\
\rowcolor{tabelcellgrey} \multicolumn{11}{l}{\textbf{\textit{Closed-Source MLLMs}}} \\
Gemini-1.5-Pro & 78.6/66.2/71.9 &  80.1/66.6/72.7 &  80.4/65.6/72.2 &  77.5/65.9/71.2 &  81.4/64.2/71.8 &  82.1/63.4/71.5 &  81.8/67.0/73.7 &  79.7/65.6/72.0 &  79.7/67.0/72.8 &  81.0/66.2/72.9   \\
GPT-4o & 83.6/53.0/64.9 &  84.0/57.1/68.0 &  86.0/59.1/70.0 &  85.6/57.3/68.7 &  84.5/58.4/69.1 &  83.7/60.2/70.0 &  82.3/64.1/72.1 &  82.8/59.2/69.0 &  85.2/61.4/71.4 &  83.0/61.2/70.5    \\

        \bottomrule
    \end{tabularx}
    \caption{Performance on MCTS-VCB across different video types. Each cell contains ``\textbf{Precision} / \textbf{Recall} / \textbf{F1 Score}\textquotedblright}
    \label{tab: all_result_video_category}
\end{minipage}}
\end{table}

\clearpage
\begin{table}
\footnotesize
\rotatebox{90}{\begin{minipage}[b][\textwidth][c]{\textheight}
\centering
    \begin{tabularx}{\textheight}{ 
			>{\raggedright\arraybackslash\hsize=2.1\hsize\linewidth=\hsize}X
			>{\centering\arraybackslash\hsize=0.9\hsize\linewidth=\hsize}X
			>{\centering\arraybackslash\hsize=0.9\hsize\linewidth=\hsize}X
			>{\centering\arraybackslash\hsize=0.9\hsize\linewidth=\hsize}X
			>{\centering\arraybackslash\hsize=0.9\hsize\linewidth=\hsize}X
			>{\centering\arraybackslash\hsize=0.9\hsize\linewidth=\hsize}X
			>{\centering\arraybackslash\hsize=0.9\hsize\linewidth=\hsize}X
			>{\centering\arraybackslash\hsize=0.9\hsize\linewidth=\hsize}X
			>{\centering\arraybackslash\hsize=0.9\hsize\linewidth=\hsize}X
			>{\centering\arraybackslash\hsize=0.9\hsize\linewidth=\hsize}X
			>{\centering\arraybackslash\hsize=0.9\hsize\linewidth=\hsize}X
			>{\centering\arraybackslash\hsize=0.9\hsize\linewidth=\hsize}X
		}
        \toprule
        {Model Name} & 0.70 & 0.72 & 0.74 & 0.76 & 0.78 & 0.80 & 0.82 & 0.84 & 0.86 & 0.88 & 0.90 \\
        \midrule
\rowcolor{tabelcellgrey} \multicolumn{12}{l}{\textbf{\textit{Open-Source MLLMs}}} \\
InternVL2\_5-8B & 68.2/37.2/48.1 &  68.5/37.8/48.8 &  68.8/38.4/49.3 &  69.8/38.9/50.0 &  69.1/39.6/50.3 &  69.6/40.1/50.8 &  69.4/40.6/51.2 &  69.7/41.0/51.7 &  69.3/41.5/51.9 &  70.1/41.9/52.4 &  69.0/42.3/52.5   \\
InternVL2\_5-26B & 68.7/37.9/48.8 &  69.1/38.6/49.5 &  70.1/39.2/50.3 &  69.5/39.7/50.6 &  69.9/40.4/51.2 &  70.1/40.9/51.7 &  70.4/41.4/52.1 &  70.3/41.8/52.5 &  70.6/42.3/52.9 &  70.2/42.7/53.1 &  69.7/43.2/53.3   \\
InternVL2\_5-38B & 68.8/38.1/49.0 &  69.5/38.8/49.8 &  70.6/39.4/50.6 &  70.4/39.9/51.0 &  70.7/40.5/51.5 &  70.6/41.0/51.8 &  70.2/41.5/52.1 &  70.5/41.9/52.6 &  70.0/42.4/52.8 &  70.3/42.8/53.2 &  70.3/43.2/53.6   \\
InternVL2\_5-78B & 68.9/38.5/49.4 &  69.8/39.4/50.3 &  70.0/39.9/50.8 &  70.2/40.5/51.4 &  70.5/41.1/51.9 &  70.6/41.6/52.4 &  70.9/42.1/52.9 &  70.4/42.6/53.1 &  70.3/43.1/53.4 &  70.5/43.5/53.8 &  70.7/43.9/54.2   \\
LLaVa\_OV\_Qwen2\_7B & 77.6/48.6/59.8 &  78.1/49.4/60.5 &  78.3/50.0/61.0 &  78.8/50.6/61.6 &  79.4/51.3/62.3 &  79.6/51.8/62.8 &  79.5/52.4/63.1 &  79.9/52.8/63.6 &  79.8/53.3/63.9 &  79.9/53.8/64.3 &  79.6/54.2/64.5   \\
LLaVa\_OV\_Qwen2\_72B & 80.0/49.8/61.4 &  80.1/50.5/61.9 &  80.8/51.1/62.6 &  80.7/51.7/63.0 &  81.5/52.4/63.8 &  81.3/52.9/64.1 &  81.8/53.4/64.7 &  81.7/54.0/65.0 &  82.1/54.4/65.5 &  81.4/54.9/65.6 &  81.7/55.3/66.0   \\
LLaVa\_Video\_Qwen2\_7B & 77.9/45.7/57.6 &  78.8/46.4/58.4 &  79.2/47.0/59.0 &  79.4/47.7/59.6 &  79.6/48.4/60.2 &  79.7/48.9/60.6 &  80.3/49.5/61.2 &  80.0/50.0/61.5 &  79.0/50.5/61.6 &  79.6/50.9/62.1 &  79.8/51.4/62.5   \\
LLaVa\_Video\_Qwen2\_72B & 80.0/48.4/60.3 &  80.7/49.2/61.1 &  80.9/49.8/61.7 &  81.5/50.5/62.3 &  81.2/51.1/62.7 &  82.1/51.7/63.4 &  81.7/52.2/63.7 &  81.7/52.7/64.1 &  82.0/53.2/64.5 &  81.7/53.7/64.8 &  81.3/54.1/65.0   \\
MiniCPM-V-2\_6 & 76.2/47.4/58.4 &  76.8/48.1/59.1 &  77.4/48.7/59.8 &  77.9/49.3/60.4 &  77.9/49.9/60.9 &  77.7/50.4/61.2 &  78.2/51.0/61.7 &  78.3/51.4/62.1 &  78.6/51.8/62.5 &  78.2/52.3/62.7 &  78.2/52.6/62.9   \\
PLLaVA-7B & 54.2/28.2/37.1 &  54.4/28.8/37.6 &  54.7/29.2/38.1 &  54.9/29.7/38.5 &  54.8/30.1/38.9 &  55.1/30.4/39.2 &  55.3/30.8/39.5 &  55.0/31.1/39.7 &  54.4/31.4/39.8 &  54.0/31.7/39.9 &  55.0/31.9/40.4   \\
PLLaVA-13B & 71.2/36.0/47.8 &  71.2/36.6/48.4 &  72.1/37.1/49.0 &  72.1/37.7/49.5 &  72.1/38.2/49.9 &  72.5/38.7/50.4 &  72.2/39.1/50.7 &  71.7/39.5/50.9 &  72.1/39.9/51.4 &  72.3/40.3/51.8 &  71.4/40.7/51.8   \\
PLLaVA-34B & 73.7/35.7/48.1 &  73.8/36.4/48.8 &  74.7/36.9/49.4 &  74.7/37.5/49.9 &  75.2/38.0/50.5 &  74.9/38.5/50.9 &  75.2/39.0/51.4 &  75.0/39.4/51.7 &  75.0/39.8/52.0 &  75.6/40.2/52.5 &  74.9/40.7/52.7   \\
Qwen2-VL-7B & 75.8/47.0/58.0 &  76.3/48.0/58.9 &  77.0/48.8/59.7 &  77.1/49.5/60.3 &  77.3/50.2/60.9 &  77.8/50.9/61.5 &  77.8/51.5/62.0 &  77.4/52.1/62.3 &  78.1/52.6/62.9 &  77.7/53.2/63.1 &  77.4/53.7/63.4   \\
Qwen2-VL-72B & 75.5/48.4/59.0 &  76.5/49.3/60.0 &  76.4/50.1/60.5 &  76.8/50.8/61.2 &  77.0/51.5/61.7 &  77.7/52.1/62.4 &  77.5/52.8/62.8 &  77.2/53.3/63.1 &  77.3/53.8/63.5 &  77.4/54.4/63.9 &  77.7/54.9/64.3   \\
Tarsier-7b & 75.4/42.3/54.2 &  75.8/42.9/54.8 &  76.1/43.5/55.4 &  76.5/44.1/55.9 &  76.4/44.7/56.4 &  76.7/45.2/56.9 &  76.6/45.7/57.3 &  76.8/46.2/57.7 &  77.0/46.7/58.1 &  76.5/47.1/58.3 &  76.5/47.6/58.7   \\
Tarsier-34b & 77.6/42.6/55.0 &  77.7/43.3/55.6 &  77.8/43.9/56.1 &  78.5/44.5/56.8 &  79.4/45.1/57.5 &  79.6/45.7/58.0 &  79.6/46.2/58.5 &  79.3/46.7/58.8 &  78.9/47.2/59.0 &  79.6/47.6/59.6 &  79.5/48.1/59.9   \\
Llama-3-VILA1.5-8b & 74.1/40.9/52.7 &  75.8/41.7/53.8 &  75.6/42.3/54.3 &  76.1/42.9/54.9 &  75.8/43.6/55.4 &  75.6/44.2/55.8 &  75.6/44.8/56.3 &  75.9/45.3/56.7 &  76.4/45.8/57.3 &  75.6/46.3/57.4 &  75.9/46.8/57.9   \\
VILA1.5-13b & 72.9/41.0/52.5 &  73.3/41.8/53.2 &  74.0/42.4/53.9 &  75.0/43.1/54.7 &  74.8/43.8/55.2 &  73.9/44.4/55.4 &  75.1/44.9/56.2 &  74.8/45.5/56.6 &  75.4/46.0/57.1 &  75.1/46.5/57.4 &  74.9/47.0/57.7   \\
VILA1.5-40b & 76.3/44.2/56.0 &  76.6/45.0/56.7 &  77.6/45.7/57.5 &  77.7/46.4/58.1 &  78.0/47.0/58.7 &  78.3/47.6/59.2 &  77.9/48.2/59.6 &  78.2/48.7/60.1 &  78.3/49.2/60.5 &  78.9/49.8/61.0 &  78.6/50.3/61.3   \\
\rowcolor{tabelcellgrey} \multicolumn{11}{l}{\textbf{\textit{Closed-Source MLLMs}}} \\
Gemini-1.5-Pro & 78.1/61.9/69.1 &  78.7/62.2/69.5 &  79.4/62.5/70.0 &  80.4/62.9/70.6 &  80.5/63.3/70.8 &  80.9/63.6/71.2 &  81.2/64.0/71.6 &  81.1/64.4/71.8 &  81.1/64.8/72.0 &  81.1/65.1/72.2 &  81.0/65.5/72.4   \\
GPT-4o & 81.7/54.2/65.2 &  82.5/54.9/65.9 &  83.2/55.5/66.6 &  83.6/56.2/67.2 &  84.1/56.9/67.9 &  83.8/57.5/68.2 &  84.2/58.1/68.8 &  84.4/58.7/69.2 &  84.5/59.2/69.6 &  84.1/59.7/69.8 &  84.5/60.2/70.3   \\
        \bottomrule
    \end{tabularx}
    \caption{Performance on MCTS-VCB across different similarity threshold used in merging key points from different nodes \(s\). Each cell contains ``\textbf{Precision} / \textbf{Recall} / \textbf{F1 Score}\textquotedblright}
    \label{tab: all_result_threshold}
\end{minipage}}
\end{table}

\clearpage

\end{document}